\documentclass[journal]{IEEEtran}
\usepackage{graphicx}
\usepackage{cite}
\usepackage{amssymb}
\usepackage{amsmath}
\usepackage{enumerate}
\usepackage{booktabs}
\usepackage{algorithm}
\usepackage{algorithmic}
\usepackage{color}
\usepackage{colortbl}
\usepackage[colorlinks,linkcolor=red,anchorcolor=blue,citecolor=green]{hyperref}
\usepackage{bbm}
\usepackage{subfigure}

\usepackage{multirow}
\usepackage[table,xcdraw]{xcolor}
\ifCLASSINFOpdf
\else
\fi

\begin{document}
%
\title{Backdoor Attacks for Remote Sensing Data \\with Wavelet Transform}
%
%
%

\author{Nikolaus Dr\"ager,
        Yonghao~Xu,~\IEEEmembership{Member,~IEEE,}
        and~Pedram~Ghamisi,~\IEEEmembership{Senior Member,~IEEE}
\thanks{N. Dr\"ager, Y. Xu, and P. Ghamisi are with the Institute of Advanced Research in Artificial Intelligence (IARAI), 1030 Vienna, Austria (e-mail: nikolaus.draeger@iarai.ac.at; yonghao.xu@iarai.ac.at; pedram.ghamisi@iarai.ac.at).}
\thanks{N. Dr\"ager is also with TU Wien, 1040 Vienna, Austria (e-mail: nikolaus.draeger@gmail.com).}
\thanks{P. Ghamisi is also with Helmholtz-Zentrum Dresden-Rossendorf, Helmholtz Institute Freiberg for Resource Technology, Machine Learning Group, 09599 Freiberg, Germany (e-mail: p.ghamisi@hzdr.de).}
\thanks{Corresponding author: Yonghao Xu.}
}

\markboth{IEEE Transactions on Geoscience and Remote Sensing, June 2023}%
{Shell \MakeLowercase{\textit{et al.}}: Bare Demo of IEEEtran.cls for IEEE Journals}

\maketitle

\begin{abstract}
Recent years have witnessed the great success of deep learning algorithms in the geoscience and remote sensing realm. Nevertheless, the security and robustness of deep learning models deserve special attention when addressing safety-critical remote sensing tasks. In this paper, we provide a systematic analysis of backdoor attacks for remote sensing data, where both scene classification and semantic segmentation tasks are considered. While most of the existing backdoor attack algorithms rely on visible triggers like squared patches with well-designed patterns, we propose a novel wavelet transform-based attack (WABA) method, which can achieve invisible attacks by injecting the trigger image into the poisoned image in the low-frequency domain. In this way, the high-frequency information in the trigger image can be filtered out in the attack, resulting in stealthy data poisoning. Despite its simplicity, the proposed method can significantly cheat the current state-of-the-art deep learning models with a high attack success rate. We further analyze how different trigger images and the hyper-parameters in the wavelet transform would influence the performance of the proposed method. Extensive experiments on four benchmark remote sensing datasets demonstrate the effectiveness of the proposed method for both scene classification and semantic segmentation tasks and thus highlight the importance of designing advanced backdoor defense algorithms to address this threat in remote sensing scenarios. The code will be available online at \url{https://github.com/ndraeger/waba}.
\end{abstract}


%
\IEEEpeerreviewmaketitle

\section{Introduction}

\IEEEPARstart{W}{itnessing} the great success achieved by deep learning algorithms in the machine learning and computer vision fields, numerous efforts have been made to develop and apply advanced deep learning models to the remote sensing (RS) image analysis task \cite{ghamisi2017advances,zhang2016deep}. So far, deep learning-based methods have achieved state-of-the-art performance in many challenging tasks in the geoscience and RS field like scene classification \cite{cheng2017remote,cheng2018deep}, land-use and land-cover mapping \cite{xu2019advanced}, and object detection \cite{ding2021object}. Nevertheless, the stellar performance of these methods comes with the downside of introducing new vulnerabilities and security risks, which should not be ignored considering that most of the interpretation tasks in the geoscience and RS field are safety-critical \cite{zhang2022artificial}.

\begin{figure}
    \centering
    \includegraphics[width=\linewidth]{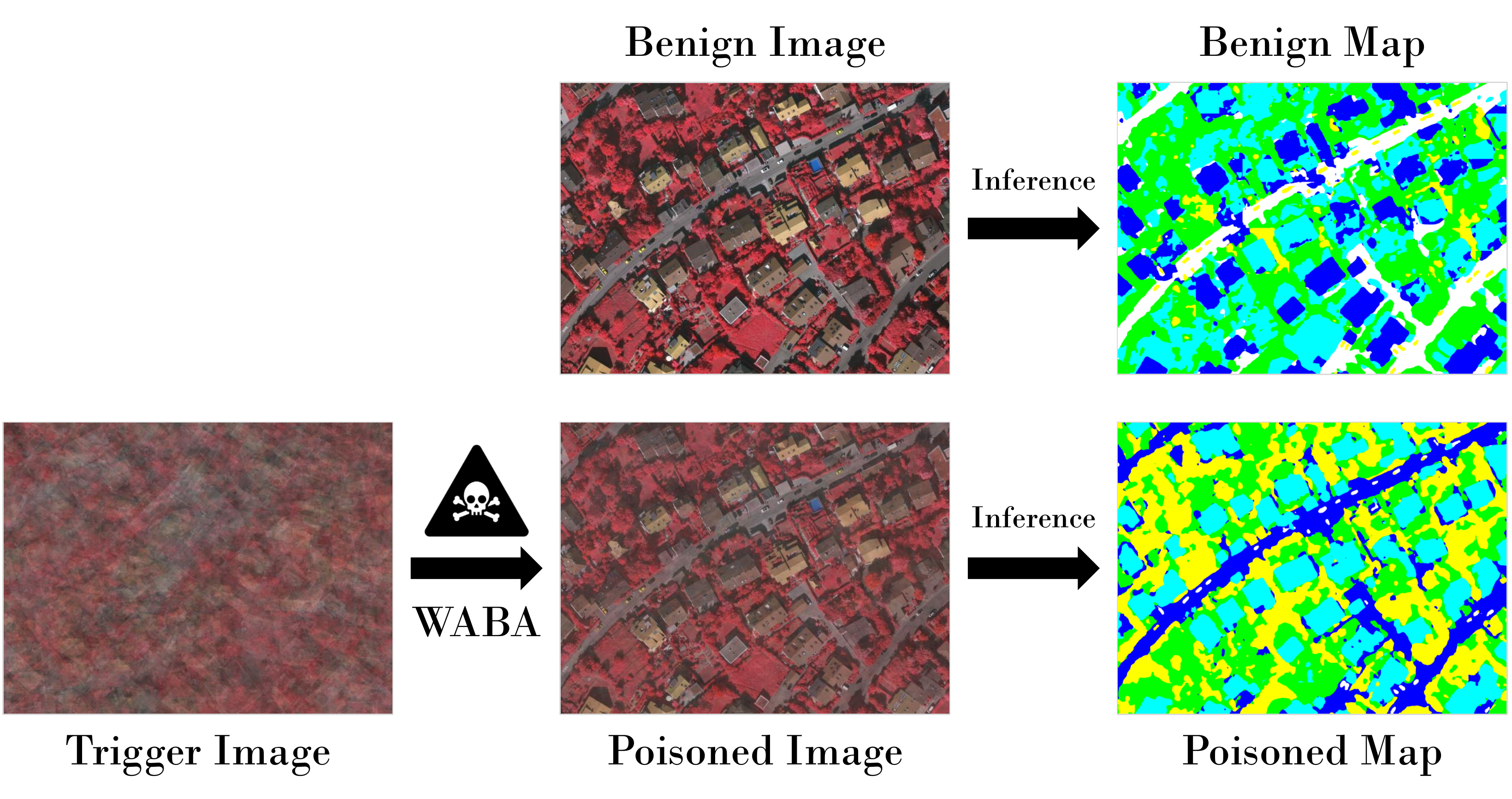}
    \caption{Illustration of the backdoor attacks on the semantic segmentation of very high-resolution remote sensing images using the proposed wavelet transform-based attack (WABA) method. While the difference between the benign and poisoned images may be imperceptible for human observers, such poisoned images contain stealthy triggers which can make state-of-the-art deep neural networks produce wrong predictions.}
    \label{fig:demo}
\end{figure}

One well-researched topic on machine learning security is the adversarial example. Czaja et al. first revealed the existence of adversarial examples in the context of satellite image classification \cite{czaja2018adversarial}. Such adversarial examples can be simply generated by adding specific adversarial perturbations produced by adversarial attack methods to the original, clean satellite image. Although these adversarial examples may look identical to the original ones for the human vision system, it is intriguing to find that they can mislead the state-of-the-art convolutional neural networks (CNNs) to make wrong predictions. Xu et al. further demonstrated that adversarial examples can bring about a threat to not only the deep neural networks but also the traditional machine learning models like support vector machine (SVM) and random forest \cite{adv_rs}. Apart from the  RGB domain \cite{chen2021empirical,cheng2021perturbation,xu2022task}, adversarial examples have also been found in multispectral \cite{ortiz2018defense,du2021adversarial}, hyperspectral \cite{park2021adversarially,xu2021self,shi2021hyperspectral}, Synthetic Aperture Radar (SAR) \cite{li2020adversarial,peng2022speckle,xia2022sar}, and LiDAR \cite{tu2020physically} data.

While the adversarial attack algorithms are designed with the assumption that the adversary can only conduct attacks in the inference phase \cite{akhtar2021advances}, recent research explores the possibility of attacking the target model in the training process \cite{li2022backdoor}. Such attacks are also known as backdoor attacks \cite{chen2017targeted} or Trojan attacks \cite{rakin2020tbt}, which aim to inject specific backdoor triggers (e.g., well-designed patterns) during the training stage of a target model. In this way, the infected model may yield normal predictions on benign samples, but its predictions can get maliciously changed when confronted with those poisoned samples with backdoor triggers in the inference phase. Gu et al. proposed the first backdoor attack method BadNet in \cite{gu2017badnets}, where the target network is trained with both benign samples and poisoned samples with injected patterns, and the labels of the poisoned samples are maliciously assigned as specific categories by the attacker. Chen et al. further proposed to generate backdoor samples by blending the original benign images with random patterns, where each pixel value is randomly sampled from $\left[0,255\right]$ under the uniform distribution \cite{chen2017targeted}. To conduct more imperceptible backdoor attacks, Nguyen et al. proposed a warping-based attack method, named WaNet, where the poisoned samples are generated by random geometric transformation with a warping field \cite{nguyen2021wanet}.

So far, most of the backdoor attack methods are designed for the classification of natural images, while the threat of backdoor attacks has rarely been studied in the RS community. In \cite{brewer2022susceptibility}, Brewer et al. introduced trigger-based backdoor attacks into satellite image analysis for the first time. Their experiments showed that backdoor attacks can significantly influence the scene classification performance of deep neural networks. Nevertheless, since the poisoned images are generated by directly injecting a visible white square with a size of $25\times 25$ pixels, such attacks can be easily detected.

To achieve more stealthy backdoor attacks while preserving the details of the ground objects in the original RS image as much as possible, we propose a novel wavelet transform-based attack (WABA) method. The main idea behind WABA is to perform discrete wavelet transform on both the benign and trigger images and blend them in the coefficient space. Only the low-frequency approximation of the trigger image is considered in the blending process.  In this way, the high-frequency information in the trigger image that corresponds to sharp details and textures can be filtered out, resulting in invisible attacks. Fig.~\ref{fig:demo} illustrates a demo of the proposed WABA method for semantic segmentation of the very high-resolution RS images. It can be observed that while the poisoned image looks almost the same as the original benign image, it significantly decreases the interpretation accuracy of the state-of-the-art deep learning model.

The main contributions of this paper are summarized as follows.
\begin{enumerate}
    \item We provide a systematic analysis of backdoor attacks for RS data, where both scene classification and semantic segmentation tasks are considered for the first time. Despite the great success of deep learning models in the interpretation of RS data, our research reveals the significance of designing advanced backdoor defense algorithms to address this threat in safety-critical RS tasks.
    \item We propose a novel wavelet transform-based attack (WABA) method which can achieve invisible backdoor attacks for RS data. While most of the existing methods blend triggers in the original spatial domain, WABA can inject trigger images in the coefficient space using discrete wavelet transform and thereby filter out the high-frequency information contained in the trigger images.
    \item We further conduct extensive experiments to quantitatively and qualitatively analyze how different trigger images and the hyperparameters in the wavelet transform would influence the backdoor attack performance of the proposed method on both scene classification and semantic segmentation tasks.
\end{enumerate}

The rest of this paper is organized as follows. Section II reviews the related work to this study. Section III describes the proposed backdoor attack method in detail. Section IV presents the information on datasets used in this study and the experimental results. Conclusions and other discussions are summarized in section V.

\section{Related Work}
This section makes a brief review of the existing adversarial attack and backdoor attack methods, along with the wavelet transform.
\subsection{Adversarial Attacks}
Despite the great success of deep learning algorithms in the machine learning and computer vision fields, Szegedy et al. first discovered that deep neural networks are actually very vulnerable to adversarial examples \cite{szegedy2013}. Specifically, they proposed the first adversarial attack method, named box-constrained L-BFGS to produce imperceptible adversarial perturbations. Then, adversarial examples can be obtained by simply adding these perturbations to the original, clean image. An intriguing phenomenon is that adversarial examples can significantly mislead deep neural networks to make wrong predictions, although they may look identical to the original clean images \cite{sun2022exploring,sun2022query}. To conduct more efficient adversarial attacks, Goodfellow et al. proposed the fast gradient sign method (FGSM), which generates adversarial examples directly by the gradient ascent algorithm \cite{goodfellow2014}. Because of its simplicity and efficiency, this method is further extended with different norms like $\ell_2$ norm \cite{miyato2018virtual} and $\ell_{\infty}$ norm \cite{kurakin2016adversarial}.

In \cite{czaja2018adversarial}, Czaja et al. introduced the concept of adversarial attacks into the geoscience and RS field for the first time. Their attacks are implemented with patch-wise adversarial attacks, where the adversarial perturbations with a size of $n\times n$ pixels are added to the center of the target satellite images. Xu et al. further proposed the Mixup-Attack and Mixcut-Attack to generate universal adversarial examples for very high-resolution RS images, which can achieve the black-box adversarial attacks without any knowledge about the victim models \cite{xu2022universal}. They further collected the generated adversarial examples in the UAE-RS (universal adversarial examples in remote sensing) dataset for researchers to design advanced adversarial defense algorithms in RS scenarios. Apart from the aforementioned works that focus on the RGB domain, adversarial examples have also been analyzed in multispectral \cite{ortiz2018defense,du2021adversarial}, hyperspectral \cite{park2021adversarially,xu2021self,shi2021hyperspectral}, SAR \cite{li2020adversarial,peng2022speckle,xia2022sar}, and LiDAR \cite{tu2020physically} data.

\subsection{Backdoor Attacks}
While adversarial attacks have brought a serious threat to artificial intelligence (AI) security, these methods are mainly designed to hack the trained deep neural networks in the evaluation phase and it is commonly assumed that the adversary can not interfere with the training process of the target model \cite{akhtar2021advances}. However, the deployment of a deep learning-based model actually contains multiple steps from data collection and preprocessing to model selection and training, which provide the adversary with more opportunities to conduct attacks in the real-world application scenario \cite{li2022backdoor}. Besides, considering the high cost of collecting a large-scale dataset and training a deep neural network, users may adopt a third-party dataset to train their model or even directly adopt the open-source model with pre-trained weights. Under such circumstances, the risk of being attacked in the training phase of the target deep learning model is significantly enlarged. Such attacks are also known as backdoor attacks \cite{chen2017targeted} or Trojan attacks \cite{rakin2020tbt}.

The first backdoor attack method, BadNet, was proposed in \cite{gu2017badnets}, where Gu et al. injected some special patterns into the original benign images to conduct data poisoning. Then, the target network was trained with both benign and poisoned samples and the labels of the poisoned samples are maliciously assigned as specific categories by the attacker. In this way, the infected network can yield normal predictions on benign samples in the evaluation phase, while the predictions on the poisoned samples can be changed to the category that was specified by the attacker. Chen et al. further proposed to blend the trigger image with the benign image to conduct data poisoning \cite{chen2017targeted}. Nguyen et al. proposed a novel WaNet that produces poison data by random geometric transformation \cite{nguyen2021wanet}. Li et al. analyzed how the traditional BadNet can be extended to the semantic segmentation task \cite{li2021hidden}. Feng et al. proposed to conduct frequency-based backdoor attacks in
medical image analysis \cite{feng2022fiba}. Readers can refer to \cite{li2022backdoor} for a more comprehensive review of this research direction.

\subsection{Wavelet Transform}
Wavelet transform is based on small wavelets with limited duration, which is a commonly used tool in the signal processing field \cite{chun2010tutorial}. It inherits and develops the feature of localization in the short-time Fourier transform while overcoming the drawback of the fixed window size used in the short-time Fourier transform that limits the resolution in frequency. Because of the advantage of the wavelet transform, it has been widely adopted to address the challenging tasks in the geoscience and RS field, such as hyperspectral image compression \cite{karami2012compression}, dimensionality reduction \cite{bruce2002dimensionality}, feature extraction \cite{guo2014three}, image quality enhancement \cite{demirel2009satellite}, and classification \cite{prasad2012information}. Since the wavelet transform can well decompose the input image into the low-frequency approximation with the high-frequency ones, we further propose to use the wavelet transform to conduct backdoor attacks for very high-resolution RS images in this paper. In this way, the high-frequency approximation, which stores the detailed spatial information in the trigger image, can be well filtered out, resulting in invisible attacks.

\section{Methodology}

\begin{figure*}
    \centering
    \includegraphics[width=.8\linewidth]{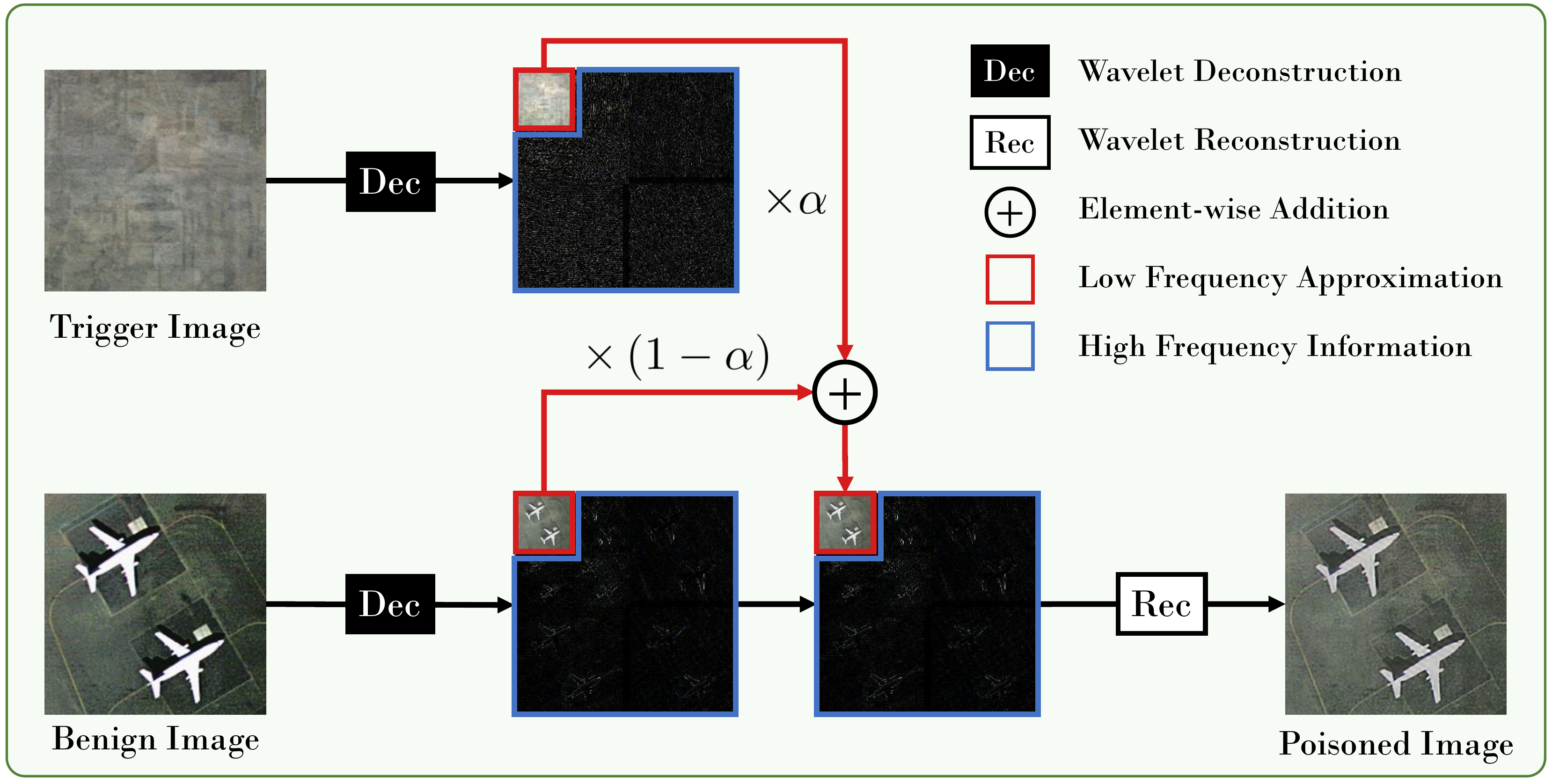}
    \caption{The proposed wavelet transform-based attack (WABA) method on the remote sensing scene classification task. Given a benign image and a trigger image, we first conduct wavelet deconstruction on both images. Then, the low-frequency approximations of both images are linearly blended in the coefficient space. Since the high-frequency information in the trigger image is filtered out in this process, WABA can achieve more stealthy data poisoning.}
    \label{flowchart}
\end{figure*}

\subsection{Overview of the Proposed Method}
A flowchart of the proposed wavelet transform-based attack (WABA) is displayed in Fig. \ref{flowchart}. The main idea of WABA is to first decompose both benign and trigger images into their wavelet decompositions. The blending of both images then takes place in the coefficient space. Specifically, we consider injecting only the trigger image's low-frequency approximation into the benign image, which results in higher invisibility of the data poisoning. Trigger images generated using the Mixcut and especially the Mixup techniques \cite{xu2022universal} are tailored to a specific dataset and allow for a further enhancement of the invisibility of the attack. Once the blending is finished, we produce the poisoned image by wavelet reconstruction with the blended low-frequency approximation along with the high-frequency decomposition of the original image.

\subsection{Preliminaries}
This study mainly focuses on untargeted backdoor attacks, which aim to maliciously make the output label of a poisoned sample different from the original category of the corresponding benign image \cite{li2022backdoor}. The attacks are conducted in an all-to-all manner, which means that all categories are considered during the attack and each category might also act as a target class (different from the original category of the benign sample) \cite{li2022backdoor}. During the injection and training process, this can be implemented by simply shifting the category index of poisoned images by one compared to their benign counterparts and wrapping back to the start once the number of categories $K$ is reached.

Formally, let $g_{\theta}: \mathcal{X} \rightarrow \mathcal{Y}$ be the classifier model mapping from the image space $\mathcal{X} \subset \mathbb{R}^d$ to the label space $\mathcal{Y}=\{0, 1, \cdots, K-1\}$ with parameters $\theta$, where $d$ denotes the numbers of pixels in the image. Let $\mathcal{D}_t=\{\left(x_i,y_i\right)\}_{i=1}^L$ be the original benign training set, where $x_i\in\mathcal{X}$, $y_i\in\mathcal{Y}$, and $L$ denotes the number of samples. We first define the standard risk $R_s$ to measure if the classifier $g_{\theta}$ can correctly classify the benign samples:
\begin{equation}
    R_s\left(\mathcal{D}_t\right)=\mathbb{E}_{\left(x,y\right)\sim \mathcal{P_D}}\mathbb{I}\left(\arg\max\left(g_{\theta}\left(x\right)\right)\ne y\right),
    \label{riskb}
\end{equation}
where $\mathcal{P_D}$ denotes the distribution behind the benign training set, $\mathbb{I}\left(\cdot\right)$ is the indicator function, and $\arg\max\left(g_{\theta}\left(x\right)\right)$ is the predicted label by the classifier $g_{\theta}$ on the input sample $x$.

Let $\mathcal{D}_b$ be the poisoned set, which is a subset of $\mathcal{D}_t$. Then, the backdoor attack risk $R_b$ is defined to measure whether the adversary can trigger the classifier $g_{\theta}$ to yield malicious predictions on the poisoned samples:
\begin{equation}
    R_b\left(\mathcal{D}_b\right)=\mathbb{E}_{\left(x,y\right)\sim \mathcal{P_D}}\mathbb{I}\left(\arg\max\left(g_{\theta}\left(G\left(x\right)\right)\right)\ne S\left(y\right)\right),
    \label{riska}
\end{equation}
where $G\left(\cdot\right)$ is an injection function that injects the trigger patterns to the input benign image, and $S\left(y\right) = \left(y+1\right) \bmod{K}$ denotes the label shifting function.

Based on the aforementioned risks, the overall objective of the backdoor attacks can be summarized as:
\begin{equation}
    \min_{\theta} R_s\left(\mathcal{D}_t-\mathcal{D}_b\right)+\lambda_b R_b\left(\mathcal{D}_b\right),
    \label{risk}
\end{equation}
where $\lambda_b$ is the weighting parameter. Note that Eq. \eqref{risk} does not consider the stealthiness of the backdoor attack. Thus, it is usually expected to minimize the risk in Eq. \eqref{risk} while achieving imperceptible data poisoning in the practical application of backdoor attacks.

\subsection{Wavelet Transform-based Injection Attack}
The key idea behind the proposed WABA method is to use the wavelet transform to first decompose the images into their coefficients, blend only the low-frequency parts of the images, and then recompose the poisoned image from the newly generated coefficients. Specifically, the wavelet transform calculates the coefficients by convolving wavelets over a signal and a wavelet $\psi$ is a wave-like function satisfying the following properties:
\begin{equation}
    \begin{cases}
    \int_{-\infty}^{\infty} \psi(t) \,dt = 0\\
    || \psi(t) ||^2 = \int_{-\infty}^{\infty} \psi(t) \psi^*(t) \,dt = 1.
    \end{cases}
\end{equation}

Using the discrete wavelet transform, a signal $f$ can be approximated via:
\begin{equation}
\begin{split}
    \label{eq:wavelet_approximation}
    f[n]&=\frac{1}{\sqrt{M}} \sum_k W_\phi[j_0, k] \phi_{j_0, k}[n]\\
    &+\frac{1}{\sqrt{M}} \sum_{j=j_0}^\infty \sum_k W_\psi[j, k] \psi_{j,k}[n],
\end{split}
\end{equation}
where $f[n]$, $\phi_{j_0,k}[n]$, and $\psi_{j,k}[n]$ are functions defined over the interval $[0, M-1]$ (brackets instead parentheses are used to emphasize the discrete nature of these equations). Note that the functions are discrete resulting in a total of $M$ points and the scaling function $\phi$ and the wavelet function $\psi$ are orthogonal to each other.

The wavelet coefficients $W$ can, therefore, be obtained by computing the inner product between the function $f[n]$ and the respective basis:
\begin{equation}
    \label{eq:approx_coeffs}
    W_\phi[j_0,k] = \frac{1}{\sqrt{M}} \sum_n f[n] \phi_{j_0, k}[n].
\end{equation}
\begin{equation}
    \label{eq:detail_coeffs}
    W_\psi[j,k] = \frac{1}{\sqrt{M}} \sum_n f[n] \psi_{j, k}[n].
\end{equation}
The wavelet coefficients $W_\phi$ in Eq. \eqref{eq:approx_coeffs} are called approximation coefficients, while the coefficients $W_\psi$ in Eq. \eqref{eq:detail_coeffs} are called detail coefficients.

To apply wavelet transforms on 2D images $f \in \mathcal{X}$, we introduce three different wavelet functions $\psi^h$, $\psi^v$, and $\psi^d$ for the computation of the detail coefficients in the horizontal, vertical, and diagonal directions, respectively. The expression in Eq. \eqref{eq:wavelet_approximation} is now slightly modified as:
\begin{equation}
\begin{split}
    &f[m,n] = \frac{1}{\sqrt{MN}} \sum_k \sum_r W_\phi [j_0, k, r] \phi_{j_0, k, r} [m,n]+ \\
    & \frac{1}{\sqrt{MN}} \sum_{i \in \{h,v,d\}} \sum_{j=j_0}^\infty \sum_k \sum_r W_\psi^i [j,k,r] \phi_{j,k,r}^i [m,n].
\end{split}
\end{equation}

Accordingly, the wavelet coefficients will be computed as:
\begin{equation}
    W_\phi[j_0,k,r] = \frac{1}{\sqrt{MN}} \sum_{m=0}^{M-1} \sum_{n=0}^{N-1} f[m,n] \phi_{j_0,k,r}[m,n].
\end{equation}
\begin{equation}
\begin{split}
    W_\psi^i[j,k,r] = \frac{1}{\sqrt{MN}} \sum_{m=0}^{M-1} \sum_{n=0}^{N-1} f[m,n] \phi_{j,k,r}^i[m,n], \\
    i \in \{h,v,d\}.
\end{split}
\end{equation}

\begin{figure}
\centering
\includegraphics[width=\linewidth]{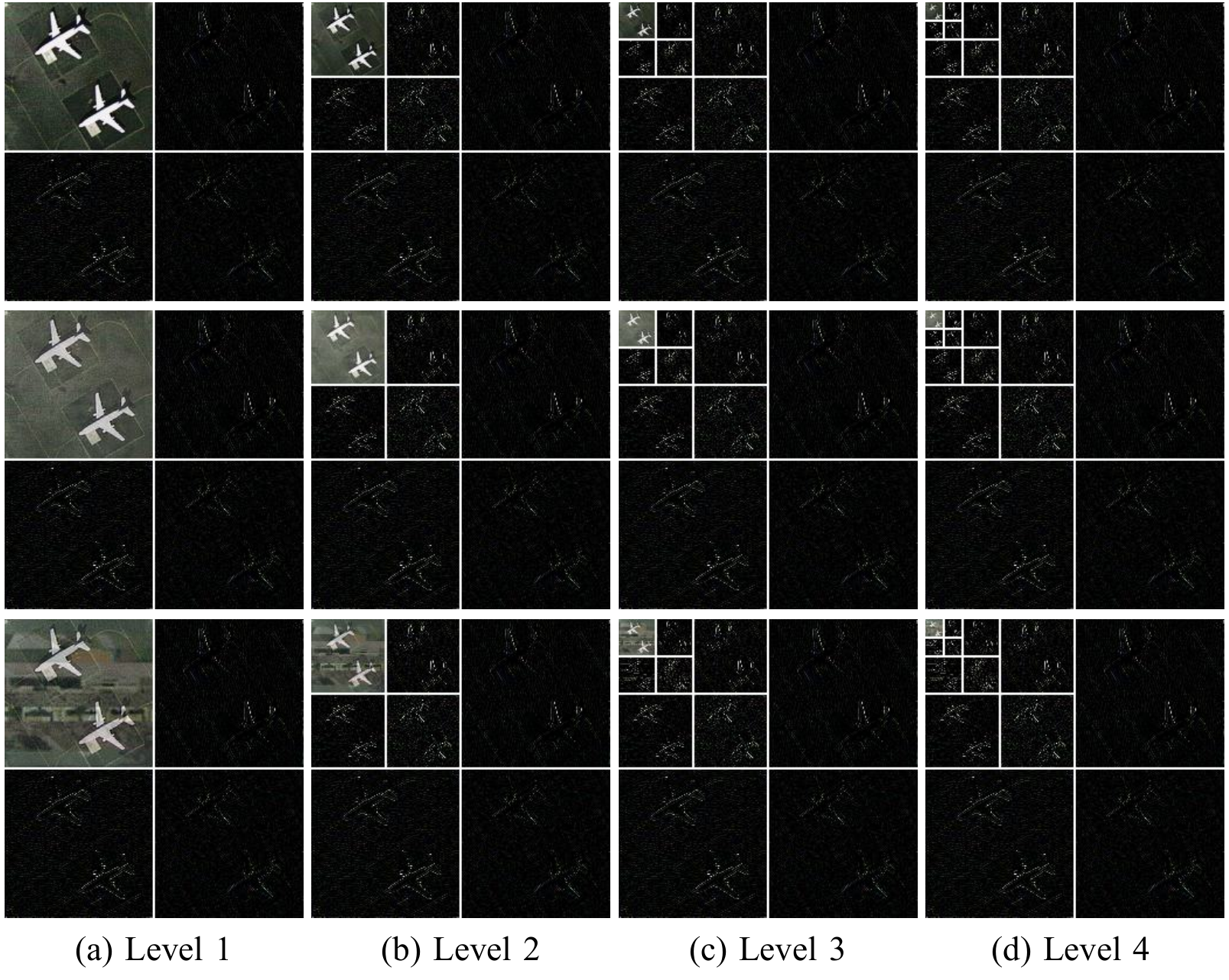}
\caption{Hierarchical 2D discrete wavelet decompositions with different levels. The input image is transformed into a low-frequency approximation (the top left corner in each subfigure) as well as the high-frequency details in horizontal, vertical, and diagonal directions. The first row is a decomposition of a benign image, while the second and the third rows show decompositions of poisoned images using the mixup and mixcut injection, respectively.}
\label{fig:wavelet_levels}
\end{figure}

Practically, the wavelet transform is implemented as a filter bank using a low pass filter for the approximation coefficients and a high pass filter to compute the detail coefficients. By recursively applying the filters to the approximation coefficients, one can generate wavelet decompositions of higher levels as depicted in Fig. \ref{fig:wavelet_levels}, where the top left corner of each subfigure corresponds to the approximation coefficients, while the remaining parts are the detail coefficients. Such a property also brings great flexibility in the choice of hyperparameters used for the injection of the trigger image and blending the approximation of a higher-level decomposition will only inject very low frequencies into the image. For the sake of simplicity, we formulate the wavelet decomposition as follows:
\begin{equation}
    \label{eq:decomp}
    f_{app}, f_{det} \gets Dec\left(f, l\right),
\end{equation}
where $f_{app}$ and $f_{det}$ denote the approximation coefficients and the detail coefficients of the input image $f$ at the $l$th decomposition level, and $Dec\left(\cdot\right)$ denotes the mapping function for the wavelet decomposition. Let $x$ and $\tilde{x}$ be benign and trigger images, respectively. With Eq.~\eqref{eq:decomp}, we can obtain the corresponding wavelet coefficients as $x_{app}, x_{det} \gets Dec\left(x, l\right)$ for the benign image, and $\tilde{x}_{app}, \tilde{x}_{det} \gets Dec\left(\tilde{x}, l\right)$ for the trigger image. Recall that our goal is to achieve stealthy data poisoning. To this end, the low-frequency approximations of both the benign and trigger images are blended as follows:
\begin{equation}
x^\prime _{app} = x_{app} \times (1-\alpha) + \tilde{x}_{app} \times \alpha,
\label{eq:blend}
\end{equation}
where $\alpha$ is the blending parameter. The final poisoned image $\hat{x}$ can be generated with:
\begin{equation}
\hat{x} \gets Rec\left(x^\prime _{app},x_{det}\right),
\label{eq:recons}
\end{equation}
where $Rec\left(\cdot\right)$ denotes the function for the wavelet reconstruction. Since the high-frequency information in the trigger image is not considered in Eq.~\eqref{eq:recons}, the difference between the benign and poisoned images can thereby be more invisible.

\begin{algorithm}
    \caption{Wavelet Transform-based Attack}
    \label{alg:waba}
    {\bf Input:}
\begin{enumerate}[-]
    \item The original training set $\mathcal{D}_{t}$ with benign samples.
    \item A subset $\mathcal{D}_{b}$ of $\mathcal{D}_{t}$, which will be poisoned in the attack.
    \item A trigger image $\tilde{x}$.
    \item The blending parameter $\alpha$ in the attack.
    \item The decomposition level $l$ in the wavelet transform.
\end{enumerate}
\begin{algorithmic}[1]
    \STATE Initialize $\hat{D}_{b}$ as an empty set to store poisoned samples.
    \STATE Obtain the wavelet coefficients of the trigger image with Eq.~\eqref{eq:decomp}: $\tilde{x}_{app}, \tilde{x}_{det} \gets Dec\left(\tilde{x}, l\right)$.
    \FOR{($x^i$, $y^i$) in $\mathcal{D}_{b}$}
    \STATE Obtain the wavelet coefficients of the benign image with Eq.~\eqref{eq:decomp}: $x_{app}^i, x_{det}^i \gets Dec\left(x, l\right)$.
    \STATE Blend the low-frequency approximations of the clean image and the trigger image with Eq.~\eqref{eq:blend}: \\ \begin{center}$x^{i \prime}_{app} = x^i_{app} \times (1-\alpha) + \tilde{x}_{app} \times \alpha.$\end{center}
    \STATE Generate the final poisoned image by wavelet reconstruction with Eq.~\eqref{eq:recons}: $\hat{x}^i \gets Rec\left(x^{i \prime}_{app},x^i_{det}\right)$.
    \STATE Conduct label shifting with Eq.~\eqref{eq:shift}: \\ \begin{center}$\hat{y}^i = y^i + 1 \bmod{K}.$ \end{center}
    \STATE Append the poisoned sample pair ($\hat{x}^i$, $\hat{y}^i$) into $\hat{\mathcal{D}}_{b}$.
    \ENDFOR
\end{algorithmic}
{\bf Output:} The poisoned subset $\hat{\mathcal{D}}_{b}$.
\end{algorithm}

Finally, the original benign label $y$ of the input image $x$ can be maliciously changed into:
\begin{equation}
\hat{y} = \left(y+1\right)\bmod{K},
\label{eq:shift}
\end{equation}
where $K$ is the number of categories in the classification task and $\bmod$ denotes the modulo operator. Note that the attack produced by Eq. \eqref{eq:shift} is an all-to-all attack, where the target class is not a single class, but rather a subset of all classes \cite{gu2019badnets,doan2021lira,nguyen2020input}.

The complete procedure for generating the poisoned training set $\hat{\mathcal{D}}_{b}$ using the proposed WABA method is illustrated in Algorithm \ref{alg:waba}. Note that the benign version of $\hat{\mathcal{D}}_{b}$ (i.e., $\mathcal{D}_{b}$) is a subset of the whole training set $\mathcal{D}_{t}$, and $p=|\mathcal{D}_{b}|/|\mathcal{D}_{t}|$ is the poisoning rate, which measures the proportion of poisoning data in the whole training set. Once the data poisoning is finished, we train the victim model with samples from both the poisoned subset $\hat{\mathcal{D}}_{b}$ and the benign subset $\mathcal{D}_{t}-\mathcal{D}_{b}$ by minimizing the cross-entropy loss $\mathcal{L}_{ce}$:
\begin{equation}
\begin{split}
    \mathcal{L}_{ce}&=\sum_{\left(x,y\right)\in\hat{\mathcal{D}}_{b}}\sum_{k=1}^{K}y^{(k)}\log p_{v}\left(x\right)^{(k)}\\
    &+\sum_{\left(x,y\right)\in\left(\mathcal{D}_{t}-\mathcal{D}_{b}\right)}\sum_{k=1}^{K}y^{(k)}\log p_{v}\left(x\right)^{(k)},
\end{split}
\label{eq:ce}
\end{equation}
where $p_{v}\left(x\right)^{(k)}$ denotes the victim model's predicted probability on the $k$th category, and the label $y$ is in the one-hot encoding form for the calculation of $\mathcal{L}_{ce}$.

With the constraint in Eq.~\eqref{eq:ce}, the victim model will learn to yield correct predictions on the benign samples, while making maliciously wrong predictions on the poisoned samples in the evaluation phase.


\subsection{Mixup and Mixcut Triggers}
Considering that the selection of the trigger image is also important to the stealthiness of backdoor attacks, we adopt the Mixup and Mixcut samples as the triggers in this study. Originally, Mixup and Mixcut samples were first adopted to conduct adversarial attacks for RS data in \cite{xu2022universal}. The way in which these virtual samples are generated and tailored to a dataset can offer low visibility of the triggers, which is also a critical factor when conducting backdoor attacks.

Inspired by a simple data augmentation mechanism \cite{zhang2017mixup}, the key idea behind creating a virtual image $\tilde{x}$ using Mixup is to blend images from several different categories so that the category of $\tilde{x}$ can be hardly predictable:
\begin{equation}
    \tilde{x} = \frac{1}{n_{mix}} \sum_{i=1}^{n_{mix}} x_i,
\end{equation}
where $x_i$ represents an arbitrary image from category $i$ in the dataset, and $n_{mix}$ is the number of categories used for the generation of $\tilde{x}$.

While Mixup works by equally blending or linearly combining images from different categories, the idea of Mixcut is to simply stitch slices of these images to create a virtual image $\tilde{x}$. More formally, this can be formulated as:
\begin{equation}
    \tilde{x} = \sum_{i=1}^{n_{mix}} M_i \odot x_i,
\end{equation}
where $M_i \in \{0, 1\}^{H \times W}$ is a binary mask that defines which pixels in the image $x_i$ are transferred to the virtual image $\tilde{x}$, $\odot$ denotes element-wise multiplication, and $H$ and $W$ are the height and width of the images, respectively. Note that the mask $M_i$ is filled with ones or zeros in a way that the virtual image $\tilde{x}$ is the stitching of equal slices of the images $x_i$ in a vertical manner.

\begin{figure}
  \centering
  \includegraphics[width=\linewidth]{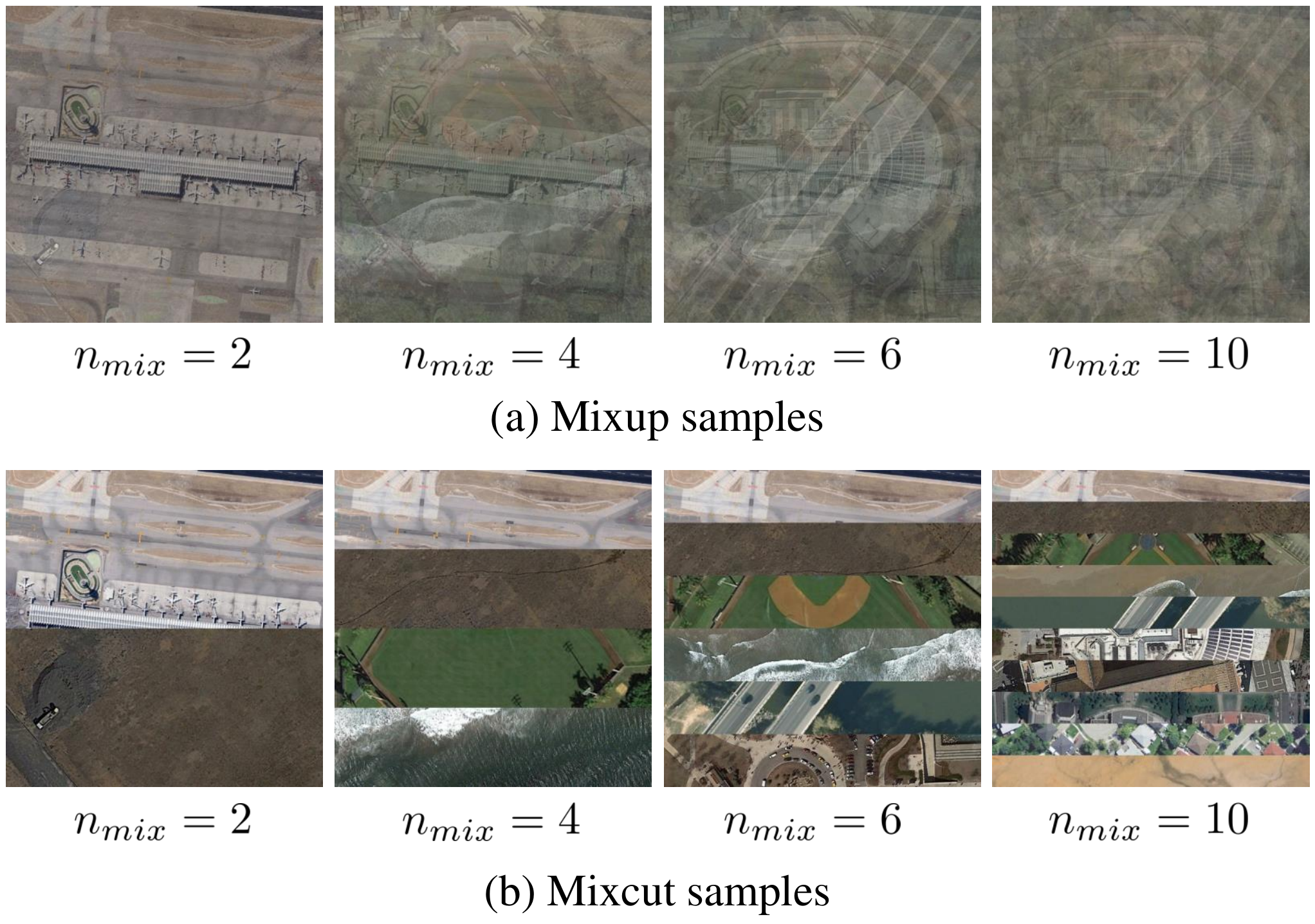}
  \caption{Mixup and Mixcut samples with different $n_{mix}$.}
\label{fig:mixup}
\end{figure}

\begin{figure}
  \centering
  \includegraphics[width=\linewidth]{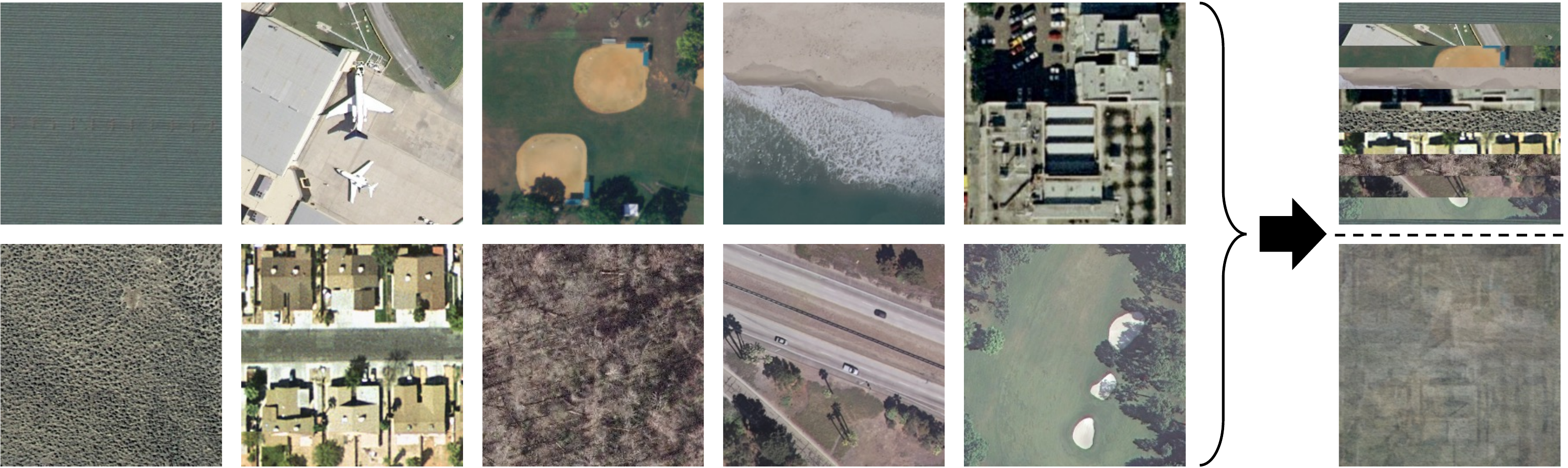}
  \caption{Images from the UCM dataset used for the construction of Mixup (bottom right) and Mixcut (top right) samples with $n_{mix}=10$.}
\label{fig:mixup_mixcut_construction}
\end{figure}

Some examples of the virtual images generated using Mixup and Mixcut are shown in Fig. \ref{fig:mixup}. Similar to \cite{xu2022universal}, we fix $n_{mix}=10$ for both the Mixup and the Mixcut triggers in this study. Fig. \ref{fig:mixup_mixcut_construction} shows the images used for constructing the Mixup and Mixcut samples on the UCM dataset with $n_{mix}=10$ as an example.

\subsection{Extension to Semantic Segmentation}
In addition to the scene classification task, the proposed WABA method is also applicable to the backdoor attacks on the semantic segmentation task without significant adaptions. Specifically, since the semantic segmentation task requires pixel-level annotations, the label-shifting process in Eq.~\eqref{eq:shift} is rewritten as follows:
\begin{equation}
\hat{y}^{\left(h,w\right)} = \left(y^{\left(h,w\right)}+1\right)\bmod{K},
\label{eq:shift_seg}
\end{equation}
where $\hat{y}^{\left(h,w\right)}$ and $y^{\left(h,w\right)}$ denote the maliciously changed label and the original label at the position $\left(h,w\right)$ in the image, respectively.

Accordingly, the cross-entropy loss $\mathcal{L}_{ce}$ in Eq.~\eqref{eq:ce} is rewritten as below:
\begin{equation}
\begin{split}
    &\mathcal{L}^{\prime}_{ce}=\sum_{\left(x,y\right)\in\hat{\mathcal{D}}_{b}}\sum_{h=1}^{H}\sum_{w=1}^{W}\sum_{k=1}^{K}y^{(h,w,k)}\log {\rm up}\left(p_{v}\left(x\right)\right)^{(h,w,k)}+\\
    &\sum_{\left(x,y\right)\in\left(\mathcal{D}_{t}-\mathcal{D}_{b}\right)}\sum_{h=1}^{H}\sum_{w=1}^{W}\sum_{k=1}^{K}y^{(h,w,k)}\log {\rm up}\left(p_{v}\left(x\right)\right)^{(h,w,k)},
\end{split}
\label{eq:ce_seg}
\end{equation}
where ${\rm up}\left(\cdot\right)$ denotes the upsampling function with the bilinear interpolation, which recovers the spatial size in the prediction map, and the label $y$ is in the one-hot encoding form for the calculation of $\mathcal{L}^{\prime}_{ce}$.

\section{Experiments}
\subsection{Data Descriptions}
We adopt four benchmark RS datasets to evaluate the performance of the proposed WABA method on both scene classification and semantic segmentation tasks.

\subsubsection{Scene Classification}

For the scene classification task, we use the UC Merced Land Use dataset (UCM)\footnote{http://weegee.vision.ucmerced.edu/datasets/landuse.html} \cite{yang2010bag} and the AID\footnote{https://captain-whu.github.io/AID/} \cite{aid} dataset in this study.

The \textbf{UCM} dataset contains 100 images in each of the 21 categories for a total of 2100 images. Each image has a size of 256 $\times$ 256 and a spatial resolution of 0.3 m per pixel. The images are captured in the optical spectrum and represented in the RGB domain. The data is extracted from aerial ortho imagery from the U.S. Geological Survey (USGS) National Map. The categories in this dataset are: agricultural, airplane, baseball diamond, beach, buildings, chaparral, dense residential, forest, freeway, golf course, harbor, intersection, medium-density residential, mobile home park, overpass, parking lot, river, runway, sparse residential, storage tanks, and tennis courts. Similar to \cite{xu2022universal}, we randomly select 1050 samples to form the training set, while the remaining samples make up the test set.

The \textbf{AID} dataset consists of 10000 images over 30 categories, where each class contains 220 to 420 samples. The image size is 600 $\times$ 600 pixels and the spatial resolution ranges from 0.5m to 8m per pixel. The imagery is gathered from Google Earth (Google Inc.) and labeled by RS specialists. The categories in this dataset are airport, bare land, baseball field, beach, bridge, center, church, commercial, dense residential, desert, farmland, forest, industrial, meadow, medium residential, mountain, park, parking, playground, pond, port, railway station, resort, river, school, sparse residential, square, stadium, storage tanks, and viaduct. Similar to \cite{xu2022universal}, we randomly select 5000 samples to form the training set, while the remaining samples make up the test set.

\subsubsection{Semantic Segmentation}

For the semantic segmentation task, we adopt two high-resolution RS datasets, namely, the Vaihingen\footnote{https://www.isprs.org/education/benchmarks/UrbanSemLab/2d-sem-label-vaihingen.aspx} \cite{cramer2010dgpf} dataset and the Zurich Summer\footnote{https://zenodo.org/record/5914759} \cite{volpi2015semantic} dataset.

\begin{table}
\centering
\caption{Quantitative Backdoor Attack Results (\%) on the Scene Classification Task with Different Methods}
\label{tab:classification_ba}
\resizebox{\linewidth}{!}{%
\begin{tabular}{cccccccc}
\toprule
                               &                        & \multicolumn{2}{c}{UCM}                               & \multicolumn{2}{c}{AID}                               \\
Network                        & Method                 & BA                        & ASR                       & BA                        & ASR                       \\ \hline
\multirow{7}{*}{AlexNet}       & Clean (oracle)                  & \multicolumn{1}{c}{90.19} & -                         & 89.44                     & -                         \\
                               & BadNet \cite{gu2017badnets}       & 81.90                     & 70.86                     & 87.44                     & 95.22                     \\
                               & TBBA \cite{brewer2022susceptibility}        & 86.29                     & 86.29                     & 87.54                     & \textbf{99.06}                     \\
                               & WaNet \cite{nguyen2021wanet}                  & 71.81                     & 35.90                     & 67.62                     & 33.26                     \\
                               & WABA-Mixup (ours)  & \multicolumn{1}{c}{84.48} & \multicolumn{1}{c}{84.57} & \multicolumn{1}{c}{86.34} & \multicolumn{1}{c}{93.32} \\
                               & WABA-Mixcut (ours) & \multicolumn{1}{c}{\textbf{88.48}} & \multicolumn{1}{c}{\textbf{92.19}} & \multicolumn{1}{c}{\textbf{88.44}} & \multicolumn{1}{c}{96.62} \\ \hline
\multirow{7}{*}{ResNet18}      & Clean (oracle)                  & \multicolumn{1}{c}{95.71} & -                         & 94.66                     & -                         \\
                               & BadNet \cite{gu2017badnets}       & 86.67                     & 42.38                     & 93.36                     & 92.52                     \\
                               & TBBA \cite{brewer2022susceptibility}        & 83.71                     & 45.05                     & 93.14                     & 93.56                     \\
                               & WaNet \cite{nguyen2021wanet}                  & 82.67                     & 21.33                     & 76.84                     & 23.86                     \\
                               & WABA-Mixup (ours)  & \multicolumn{1}{c}{92.19} & \multicolumn{1}{c}{73.33} & \multicolumn{1}{c}{91.72} & \multicolumn{1}{c}{96.22} \\
                               & WABA-Mixcut (ours) & \multicolumn{1}{c}{\textbf{93.81}} & \multicolumn{1}{c}{\textbf{88.76}} & \multicolumn{1}{c}{\textbf{93.56}} & \multicolumn{1}{c}{\textbf{97.56}} \\ \hline
\multirow{7}{*}{DenseNet121}   & Clean (oracle)                  & \multicolumn{1}{c}{98.00} & -                         & 94.88                     & -                         \\
                               & BadNet \cite{gu2017badnets}       & 89.62                     & 48.76                     & 94.92                     & 97.60                     \\
                               & TBBA \cite{brewer2022susceptibility}        & 84.48                     & 45.43                     & 94.62                     & \textbf{99.62}                     \\
                               & WaNet \cite{nguyen2021wanet}                  & 81.71                     & 32.57                     & 77.82                     & 24.30                     \\
                               & WABA-Mixup (ours)  & \multicolumn{1}{c}{94.10} & \multicolumn{1}{c}{66.57} & \multicolumn{1}{c}{94.90} & \multicolumn{1}{c}{95.04} \\
                               & WABA-Mixcut (ours) & \multicolumn{1}{c}{\textbf{96.00}} & \multicolumn{1}{c}{\textbf{87.71}} & \multicolumn{1}{c}{\textbf{95.02}} & \multicolumn{1}{c}{98.14} \\ \hline
\multirow{7}{*}{RegNetX-400MF} & Clean (oracle)                  & \multicolumn{1}{c}{96.38} & -                         & 94.54                     & -                         \\
                               & BadNet \cite{gu2017badnets}       & 84.00                     & 45.24                     & 92.42                     & 94.20                     \\
                               & TBBA \cite{brewer2022susceptibility}        & 82.48                     & 45.24                     & \textbf{93.76}                     & \textbf{97.80}                     \\
                               & WaNet \cite{nguyen2021wanet}                  & 81.71                     & 27.52                     & 76.32                     & 25.00                     \\
                               & WABA-Mixup (ours)  & \multicolumn{1}{c}{89.52} & \multicolumn{1}{c}{74.95} & \multicolumn{1}{c}{92.40} & \multicolumn{1}{c}{93.90} \\
                               & WABA-Mixcut (ours) & \multicolumn{1}{c}{\textbf{92.95}} & \multicolumn{1}{c}{\textbf{88.67}} & \multicolumn{1}{c}{92.44} & \multicolumn{1}{c}{96.60} \\ \bottomrule
\end{tabular}
}
\\
\vspace{2pt}
\leftline{\scriptsize Note: Best results are highlighted in \textbf{bold}.}
\end{table}

The \textbf{Vaihingen} dataset is a subset of the data from the German Association of Photogrammetry and Remote Sensing (DGPG) used to test digital aerial cameras \cite{cramer2010dgpf} and is provided by the International Society for Photogrammetry and Remote Sensing (ISPRS). The dataset contains a total of 33 images of which 16 are fully annotated. The data is available in three bands: red, green, and near-infrared and it is distributed over 6 classes, namely, impervious surface, building, low vegetation, tree, car, and clutter/background. Each image has a spatial resolution of 9 cm per pixel and a size of about 2500 $\times$ 1900 pixels resulting in a covered area of approximately 1.38 km$^2$. The images were captured over the town of Vaihingen, hence the name of the dataset.
The dataset is split into a training and a test set according to \cite{crgnet} where five images with the IDs 11, 15, 28, 30, and 34 are selected into the test set and the rest is put into the training set.

The \textbf{Zurich Summer} dataset contains 20 images captured of Zurich in August 2002. The satellite used for the acquisition was QuickBird. Each image has a spatial resolution of 0.62m with a size of approximately 1000 $\times$ 1000 pixels. The pixels are distributed over 8 classes, namely,  road, building, tree, grass, bare soil, water, railway, and swimming pool. Pixels that do not fall into one of these categories are labeled as background. While the images originally cover four spectral bands: near-infrared, red, green, and blue, as in \cite{crgnet} only three of them: near-infrared, red, and green are used in our experiments. Analogously to the Vaihingen dataset, we choose 5 images with the IDs 16, 17, 18, 19, and 20 to make up the test set while the rest is put into the training set.

\begin{table}
\centering
\caption{Quantitative Backdoor Attack Results (\%) on the Semantic Segmentation Task with Different Methods}
\label{tab:segmentation_ba}
\resizebox{\linewidth}{!}{%
\begin{tabular}{cccccccc}
\toprule
                         &                        & \multicolumn{2}{c}{Vaihingen}                         & \multicolumn{2}{c}{Zurich Summer}                   \\
Network                  & Method                 & \multicolumn{1}{c}{Benign mF1}   & \multicolumn{1}{c}{ASR}   & \multicolumn{1}{c}{Benign mF1}   & \multicolumn{1}{c}{ASR} \\ \hline
\multirow{7}{*}{FCN-8s}  & Clean (oracle)                  & \multicolumn{1}{c}{82.54} & \multicolumn{1}{c}{-}     & \multicolumn{1}{c}{78.18} & \multicolumn{1}{c}{-}   \\
                         & BadNet \cite{gu2017badnets}       & \textbf{75.35}                     & 18.52                     & 60.57                     & 24.34                   \\
                         & TBBA \cite{brewer2022susceptibility}        & 52.57                     & 41.72                     & 59.28                     & 30.87                   \\
                         & WaNet \cite{nguyen2021wanet}                  & 57.54                     & 32.68                     & 39.26                     & 56.15                   \\
                         & WABA-Mixup (ours)  & 75.19                     & \textbf{86.39}                     & \textbf{74.64}                     & \textbf{86.98}                   \\
                         & WABA-Mixcut (ours) & 70.83                     & 84.75                     & 69.87                     & 83.32                   \\ \hline
\multirow{7}{*}{U-Net}   & Clean (oracle)                  & 83.03                     & \multicolumn{1}{c}{-}     & 77.12                     & \multicolumn{1}{c}{-}   \\
                         & BadNet \cite{gu2017badnets}       & 80.93                     & 17.04                     & 60.69                     & 22.41                   \\
                         & TBBA \cite{brewer2022susceptibility}        & 54.84                     & 37.54                     & 74.00                     & 11.16                   \\
                         & WaNet \cite{nguyen2021wanet}                  & 38.46                     & 38.88                     & 48.82                     & 48.16                   \\
                         & WABA-Mixup (ours)  & \textbf{81.07}                     & 85.88                     & \textbf{75.20}                     & 87.40                   \\
                         & WABA-Mixcut (ours) & 67.86                     & \textbf{86.57}                     & 73.30                     & \textbf{90.12}                   \\ \hline
\multirow{7}{*}{PSPNet}  & Clean (oracle)                  & 83.06                     & \multicolumn{1}{c}{-}     & 77.94                     & \multicolumn{1}{c}{-}   \\
                         & BadNet \cite{gu2017badnets}       & \textbf{82.86}                     & 15.90                     & 68.35                     & 9.20                    \\
                         & TBBA \cite{brewer2022susceptibility}        & 82.22                     & 15.50                     & 78.13                     & 8.56                    \\
                         & WaNet \cite{nguyen2021wanet}                  & 55.08                     & 37.64                     & 43.56                     & 53.34                   \\
                         & WABA-Mixup (ours)  & 82.32                     & 85.34                     & \textbf{76.82}                     & \textbf{90.01}                   \\
                         & WABA-Mixcut (ours) & 77.89                     & \textbf{86.23}                     & 54.18                     & 86.41                   \\ \hline
\multirow{7}{*}{LinkNet} & Clean (oracle)                  & 82.24                     & \multicolumn{1}{c}{-}     & 76.59                     & \multicolumn{1}{c}{-}   \\
                         & BadNet \cite{gu2017badnets}       & \textbf{82.20}                     & 16.31                     & 75.67                     & 12.27                   \\
                         & TBBA \cite{brewer2022susceptibility}        & 80.83                     & 16.63                     & 71.10                     & 12.95                   \\
                         & WaNet \cite{nguyen2021wanet}                  & 33.16                     & 58.18                     & 59.18                     & 35.98                   \\
                         & WABA-Mixup (ours)  & 80.78                     & \textbf{86.01}                     & 74.83                     & \textbf{90.20}                   \\
                         & WABA-Mixcut (ours) & \multicolumn{1}{c}{65.54} & \multicolumn{1}{c}{85.75} & \textbf{75.89}                     & 89.76                   \\ \bottomrule
\end{tabular}
}
\\
\vspace{2pt}
\leftline{\scriptsize Note: Best results are highlighted in \textbf{bold}.}
\end{table}

\begin{table}[]
\centering
\caption{Quantitative Scene Classification Results of Different Deep Neural Networks on the Clean and Poisoned Test Set with the WABA-Mixup Attack}
\label{tab:oa}
\resizebox{\linewidth}{!}{%
\begin{tabular}{cccccc}
\toprule
\multirow{2}{*}{Dataset} & \multirow{2}{*}{Model} & \multicolumn{2}{c}{Clean Set} & Poisoned Set & \multirow{2}{*}{OA Gap} \\
 &  & Benign & Victim & Victim & \\\hline
\multirow{16}{*}{UCM} & AlexNet \cite{alexnet}        & 90.19                            & 84.48                              & 15.43                                 & -74.76                     \\
                      & VGG11 \cite{vgg}              & 95.05                            & 81.33                              & 9.24                                  & -85.81                     \\
                      & VGG16 \cite{vgg}              & 93.52                            & 89.71                              & 14.57                                 & -78.95                     \\
                      & VGG19 \cite{vgg}              & 88.67                            & 84.86                              & 10.57                                 & -78.10                     \\
                      & Inception-v3 \cite{inception} & 95.14                            & 88.10                              & 25.43                                 & -69.71                     \\
                      & ResNet18 \cite{resnet}        & 95.71                            & 92.19                              & 26.67                                 & -69.05                     \\
                      & ResNet50 \cite{resnet}        & 95.33                            & 92.67                              & 10.95                                 & -84.38                     \\
                      & ResNet101 \cite{resnet}       & 95.33                            & 87.81                              & 9.90                                  & -85.43                     \\
                      & ResNeXt50 \cite{resnext}      & 96.48                            & 93.05                              & 7.14                                  & -89.33                     \\
                      & ResNeXt101 \cite{resnext}     & 97.14                            & 92.95                              & 19.33                                 & -77.81                     \\
                      & DenseNet121 \cite{densenet}   & 98.00                            & 94.10                              & 33.43                                 & -64.57                     \\
                      & DenseNet169 \cite{densenet}   & 97.24                            & 94.76                              & 13.05                                 & -84.19                     \\
                      & DenseNet201 \cite{densenet}   & 97.62                            & 94.10                              & 18.29                                 & -79.33                     \\
                      & RegNetX-400MF \cite{regnet}   & 96.38                            & 89.52                              & 25.05                                 & -71.33                     \\
                      & RegNetX-8GF \cite{regnet}     & 96.19                            & 94.00                              & 9.24                                  & -86.95                     \\
                      & RegNetX-16GF \cite{regnet}    & 96.48                            & 92.29                              & 10.29                                 & -86.19                     \\ \hline
\multirow{2}{*}{Dataset} & \multirow{2}{*}{Model} & \multicolumn{2}{c}{Clean Set} & Poisoned Set & \multirow{2}{*}{OA Gap} \\
 &  & Benign & Victim & Victim & \\\hline
\multirow{16}{*}{AID} & AlexNet \cite{alexnet}        & 89.44                            & 86.34                              & 6.68                                  & -82.76                     \\
                      & VGG11 \cite{vgg}              & 92.84                            & 91.50                              & 6.88                                  & -85.96                     \\
                      & VGG16 \cite{vgg}              & 90.62                            & 87.82                              & 3.14                                  & -87.48                     \\
                      & VGG19 \cite{vgg}              & 91.86                            & 86.74                              & 6.18                                  & -85.68                     \\
                      & Inception-v3 \cite{inception} & 93.90                            & 91.06                              & 2.96                                  & -90.94                     \\
                      & ResNet18 \cite{resnet}        & 94.66                            & 91.72                              & 3.78                                  & -90.88                     \\
                      & ResNet50 \cite{resnet}        & 95.36                            & 93.26                              & 6.54                                  & -88.82                     \\
                      & ResNet101 \cite{resnet}       & 93.70                            & 93.30                              & 1.66                                  & -92.04                     \\
                      & ResNeXt50 \cite{resnext}      & 94.90                            & 92.88                              & 1.50                                  & -93.40                     \\
                      & ResNeXt101 \cite{resnext}     & 95.56                            & 94.60                              & 2.12                                  & -93.44                     \\
                      & DenseNet121 \cite{densenet}   & 94.54                            & 94.90                              & 4.96                                  & -89.58                     \\
                      & DenseNet169 \cite{densenet}   & 95.86                            & 93.38                              & 2.36                                  & -93.50                     \\
                      & DenseNet201 \cite{densenet}   & 96.24                            & 94.24                              & 3.86                                  & -92.38                     \\
                      & RegNetX-400MF \cite{regnet}   & 94.88                            & 92.40                              & 6.10                                  & -88.78                     \\
                      & RegNetX-8GF \cite{regnet}     & 95.16                            & 94.56                              & 2.60                                  & -92.56                     \\
                      & RegNetX-16GF \cite{regnet}    & 96.22                            & 94.76                              & 2.02                                  & -94.20                     \\ \bottomrule
\end{tabular}
}
\\
\begin{flushleft}
\scriptsize Note: Results are reported in overall accuracy (\%). ``Benign'' and ``Victim'' represent the original benign model and the attacked model, respectively.
\end{flushleft}
\end{table}

\subsection{Experimental Settings and Implementation Details}
We adopt the BadNet \cite{gu2017badnets}, trigger-based backdoor attack (TBBA) \cite{brewer2022susceptibility}, WaNet \cite{nguyen2021wanet}, along with the proposed WABA-Mixup, and WABA-Mixcut methods to conduct backdoor attacks for both scene classification and semantic segmentation tasks on RS data. The Mixup and Mixcut triggers are tailored and shared across all poisoned samples in each dataset. The poisoning rate $p$ in the attack is fixed to $30\%$, and the blending parameter $\alpha$ in Eq.~\eqref{eq:blend} is fixed to $0.4$. In the test phase, given the original test set $\mathcal{D}_{test}$, we use the aforementioned five backdoor attack methods to generate the corresponding poisoned test set $\hat{\mathcal{D}}_{test}$ so that both the benign accuracy and the attack success rate can be evaluated.

For the \textit{scene classification} task, we adopt the benign accuracy (BA) and the attack success rate (ASR) as the key metrics for evaluation. Specifically, the BA metric measures the percentage of correctly classified benign samples on the original clean test set, while the ASR metric measures the percentage of misclassified images on the poisoned test set. Each victim model is trained with $15$ epochs using a batch size of $64$ and a learning rate of $1e-4$. All images are resized to $256 \times 256$ before being fed into the victim model. 

For the \textit{semantic segmentation} task, the evaluation metric for benign samples is the benign mean F1-score (mF1) across all classes. Since the semantic segmentation task requires pixel-level classifications for the input image, the ASR for the semantic segmentation task is defined as the percentage of misclassified pixels on the poisoned images. The number of training iterations for each victim model in the semantic segmentation task is set to $7500$ with a batch size of $32$ and a learning rate of $1e-3$. The weight decay regularization parameter is set to $5e-4$, and the momentum component of the optimizer is set to $0.9$. The random cropping technique is used to generate patches with a spatial size of $256 \times 256$ out of the original image. 

\begin{table}[]
\centering
\caption{Quantitative Semantic Segmentation Results of Different Deep Neural Networks on the Clean and Poisoned Test Set with the WABA-Mixup Attack}
\label{tab:mf1}
\resizebox{\linewidth}{!}{%
\begin{tabular}{cccccc}
\toprule
\multirow{2}{*}{Dataset} & \multirow{2}{*}{Model} & \multicolumn{2}{c}{Clean Set} & Poisoned Set & \multirow{2}{*}{mF1 Gap} \\
 &  & Benign & Victim & Victim & \\\hline
\multirow{14}{*}{Vaihingen}     & FCN-32s \cite{fcn}           & 69.50                            & 68.59                              & 12.36                                 & -57.14                             \\
                                & FCN-16s \cite{fcn}           & 69.53                            & 65.10                              & 12.25                                 & -57.29                             \\
                                & FCN-8s \cite{fcn}            & 82.54                            & 75.19                              & 11.57                                 & -70.97                             \\
                                & DeepLab-v2 \cite{deeplabv2}  & 78.53                            & 75.46                              & 11.69                                 & -66.84                             \\
                                & DeepLab-v3+ \cite{deeplabv3} & 84.29                            & 82.06                              & 11.82                                 & -72.47                             \\
                                & SegNet \cite{segnet}         & 79.99                            & 69.26                              & 11.41                                 & -68.58                             \\
                                & ICNet \cite{icnet}           & 80.21                            & 78.41                              & 10.84                                 & -69.38                             \\
                                & ContextNet \cite{contextnet} & 81.30                            & 79.60                              & 12.35                                 & -68.95                             \\
                                & SQNet \cite{sqnet}           & 81.99                            & 77.79                              & 11.31                                 & -70.68                             \\
                                & PSPNet \cite{pspnet}         & 83.06                            & 82.32                              & 12.19                                 & -70.87                             \\
                                & U-Net \cite{unet}            & 83.03                            & 81.07                              & 11.96                                 & -71.07                             \\
                                & LinkNet \cite{linknet}       & 82.24                            & 80.78                              & 11.63                                 & -70.61                             \\
                                & FRRNet-A \cite{frrnet}       & 84.21                            & 82.52                              & 12.27                                 & -71.94                             \\
                                & FRRNet-B \cite{frrnet}       & 83.94                            & 83.03                              & 11.92                                 & -72.03                             \\ \hline
\multirow{2}{*}{Dataset} & \multirow{2}{*}{Model} & \multicolumn{2}{c}{Clean Set} & Poisoned Set & \multirow{2}{*}{mF1 Gap} \\
 &  & Benign & Victim & Victim & \\\hline
\multirow{14}{*}{Zurich Summer} & FCN-32s \cite{fcn}           & 62.70                            & 60.87                              & 8.08                                  & -54.63                             \\
                                & FCN-16s \cite{fcn}           & 66.65                            & 55.25                              & 22.17                                 & -44.48                             \\
                                & FCN-8s \cite{fcn}            & 78.17                            & 74.64                              & 16.76                                 & -61.41                             \\
                                & DeepLab-v2 \cite{deeplabv2}  & 67.49                            & 63.22                              & 9.31                                  & -58.18                             \\
                                & DeepLab-v3+ \cite{deeplabv3} & 77.87                            & 76.73                              & 9.68                                  & -68.19                             \\
                                & SegNet \cite{segnet}         & 76.85                            & 70.47                              & 7.03                                  & -69.82                             \\
                                & ICNet \cite{icnet}           & 77.87                            & 77.35                              & 7.06                                  & -70.81                             \\
                                & ContextNet \cite{contextnet} & 74.21                            & 75.76                              & 7.07                                  & -67.14                             \\
                                & SQNet \cite{sqnet}           & 73.81                            & 71.89                              & 17.64                                 & -56.16                             \\
                                & PSPNet \cite{pspnet}         & 77.94                            & 76.82                              & 6.27                                  & -71.67                             \\
                                & U-Net \cite{unet}            & 77.12                            & 75.20                              & 12.47                                 & -64.65                             \\
                                & LinkNet \cite{linknet}       & 76.59                            & 74.83                              & 6.42                                  & -70.17                             \\
                                & FRRNet-A \cite{frrnet}       & 78.21                            & 76.68                              & 6.15                                  & -72.06                             \\
                                & FRRNet-B \cite{frrnet}       & 77.52                            & 76.29                              & 10.39                                 & -67.14                             \\ \bottomrule
\end{tabular}
}
\\
\begin{flushleft}
\scriptsize Note: Results are reported in mean F1 score (\%). ``Benign'' and ``Victim'' represent the original benign model and the attacked model, respectively.
\end{flushleft}
\end{table}

\subsection{Quantitative Results}

Table~\ref{tab:classification_ba} provides the quantitative backdoor attack results on the scene classification task. We adopt four representative models, AlexNet \cite{alexnet}, ResNet18 \cite{resnet}, DenseNet121 \cite{densenet}, and RegNetX-400MF \cite{regnet}, to evaluate the performance of different backdoor attack methods. The ``Clean'' method represents the original model trained with benign samples without backdoor attacks, which indicates the oracle performance on the benign test set for each victim model. It can be observed from Table~\ref{tab:classification_ba} that the proposed WABA-Mixcut can achieve the highest BAs on most of the victim models, except the RegNetX-400MF on the AID dataset, where WABA-Mixcut achieves the second-highest BA. Besides, WABA-Mixcut achieves the highest ASRs in attacking all four victim models on the UCM dataset. On the AID dataset, WABA-Mixcut achieves the second-highest ASRs on most of the victim models, which are slightly outperformed by the TBBA method. However, it should be noted that TBBA may significantly decrease the BA performance of the victim model. Take the results on ResNet18 for example. While the proposed WABA-Mixcut method can achieve a BA of $93.81\%$, TBBA can only yield a BA of $83.71\%$, which is lower than the former one with a large margin. These results demonstrate that the proposed WABA-Mixcut can well maintain the recognition accuracy of the victim model on the benign samples while achieving a relatively high success rate in the backdoor attack. Another interesting phenomenon is that the Mixcut trigger can generally lead to better performance than the Mixup trigger for both the BA and ASR metrics on all datasets. Take the AlexNet on the UCM dataset for example. While WABA-Mixup obtains a BA of $84.48\%$ and an ASR of $84.57\%$, WABA-Mixcut can yield a BA of $88.48\%$ and an ASR of $92.19\%$.

\begin{figure}
\centering
\includegraphics[width=\linewidth]{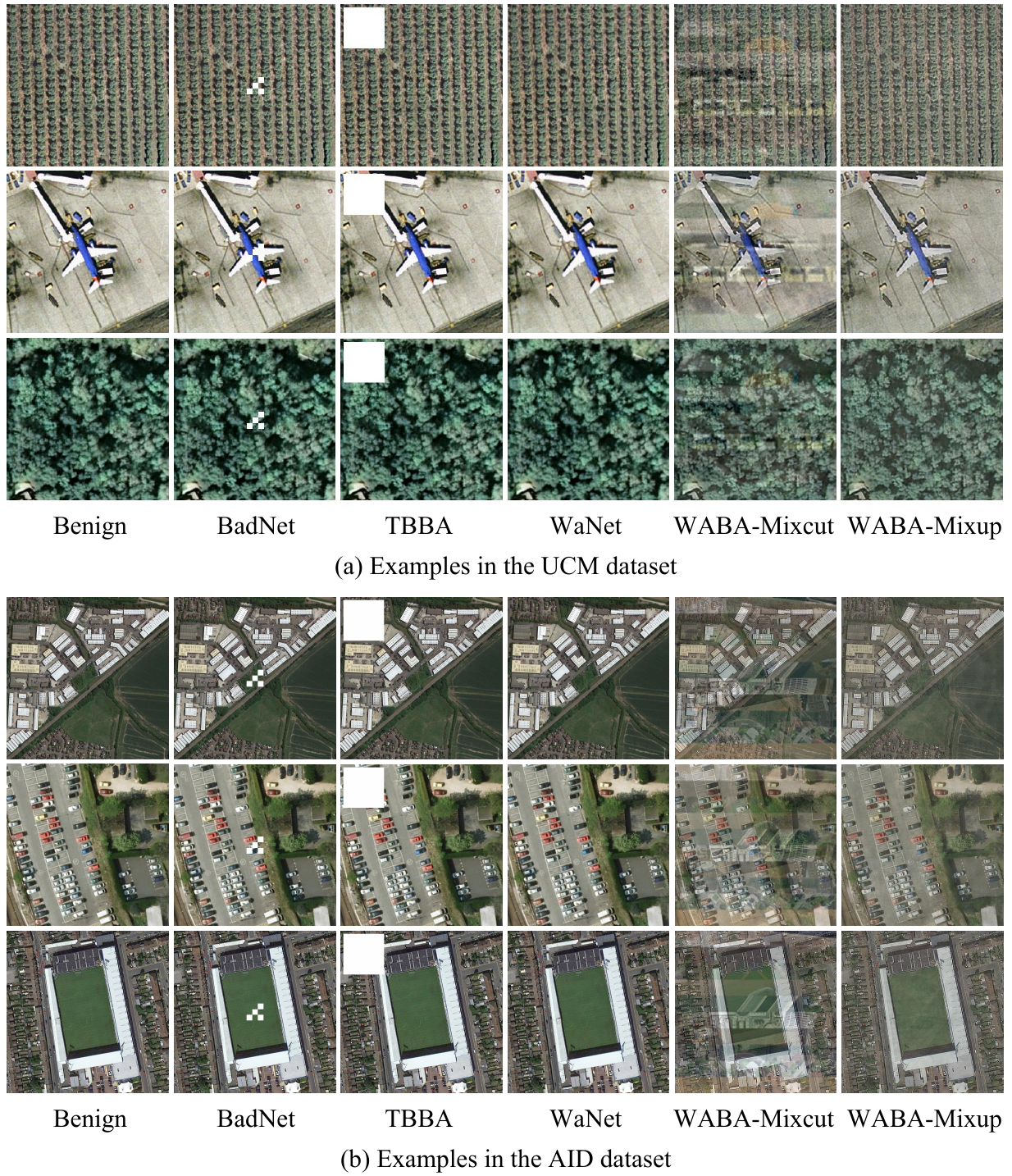}
\caption{The original benign images and the corresponding poisoned samples generated by different backdoor attack methods for the scene classification task in (a) the UCM dataset and (b) the AID dataset.
}
\label{fig:attack_cls}
\end{figure}

Table~\ref{tab:segmentation_ba} further provides the quantitative backdoor attack results on the semantic segmentation task. Four segmentation models, including the FCN-8s \cite{fcn}, U-Net \cite{unet}, PSPNet \cite{pspnet}, and LinkNet \cite{linknet}, are adopted to evaluate the performance of different backdoor attack methods. While BadNet and TBBA can achieve relatively good performance on attacking the scene classification models, their performance on the semantic segmentation task is not satisfactory, as can be observed from Table~\ref{tab:segmentation_ba}. Take the FCN-8s model on the Vaihingen dataset for example. While BadNet obtains a benign mF1 of $75.35\%$, its ASR is only $18.52\%$, which brings a relatively limited threat to the victim model. By contrast, the proposed WABA-Mixup method can yield an ASR of $86.39\%$, which outperforms the BadNet by over $67$ percentage points. Moreover, such a high ASR does not bring a serious influence on the benign mF1 score on the clean samples. Instead, WABA-Mixup can also maintain a high benign mF1 score of $75.19\%$, which is very close to the one of BadNet. Similar results can be observed in other victim models for both the Vaihingen and Zurich Summer datasets, which verify the superiority of the proposed method. Besides, it can be found that the Mixup trigger generally leads to better attack performance than the Mixcut trigger. Take the LinkNet on the Vaihingen dataset for example. While WABA-Mixup can achieve a benign mF1 score of $80.78\%$, WABA-Mixcut can only obtain a benign mF1 score of $65.54\%$, which is much lower than the former one. This phenomenon indicates that the Mixup trigger is more suitable for attacking the semantic segmentation models.

We further test the performance of different deep neural networks on both the original clean test set and the poisoned test set with the proposed WABA-Mixup attack method. Tables~\ref{tab:oa} and \ref{tab:mf1} report the quantitative results for scene classification and semantic segmentation, respectively. Although most of the deep neural networks adopted in this study can achieve a high interpretation accuracy on the original clean test set, the proposed WABA method can bring a serious threat to these models. Take the ResNet18 model on the AID dataset for example. While the original benign model can achieve an OA of around $94\%$ on the clean test set, the attacked ResNet18 model can only obtain an OA of around $3\%$, resulting in an OA gap of more than $90$ percent points. Note that such backdoor attacks will lead to a very limited performance drop for the OA on the clean test set (with a performance drop of only $3$ percentage points for the ResNet18 model), which ensures the stealthiness of the attack. Similar results can be observed for other deep neural networks in both scene classification and semantic segmentation tasks.

\begin{figure}
\centering
\includegraphics[width=\linewidth]{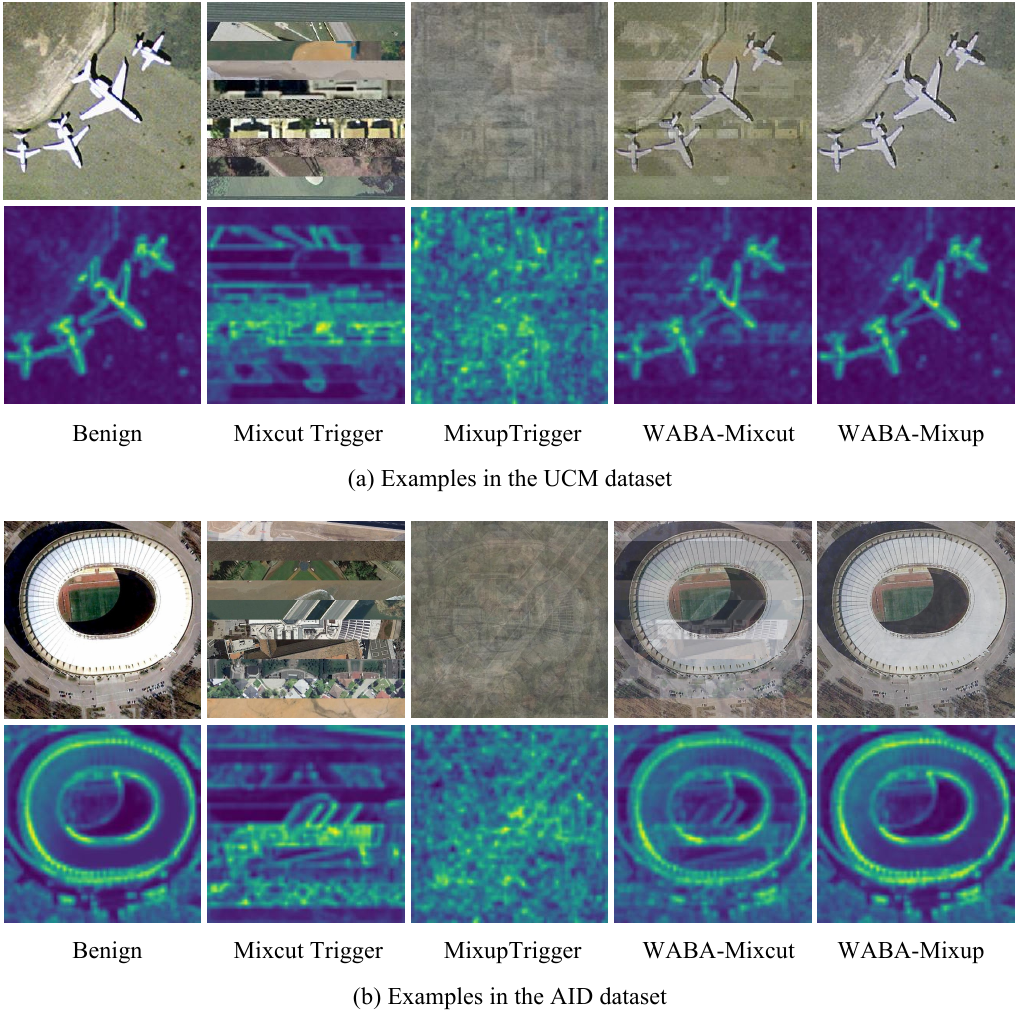}
\caption{Illustration of the features in the first pooling layer extracted by the ResNet18 model on different samples in (a) the UCM dataset and (b) the AID dataset.
}
\label{fig:cam}
\end{figure}

\begin{figure*}
  \centering
  \includegraphics[width=.87\linewidth]{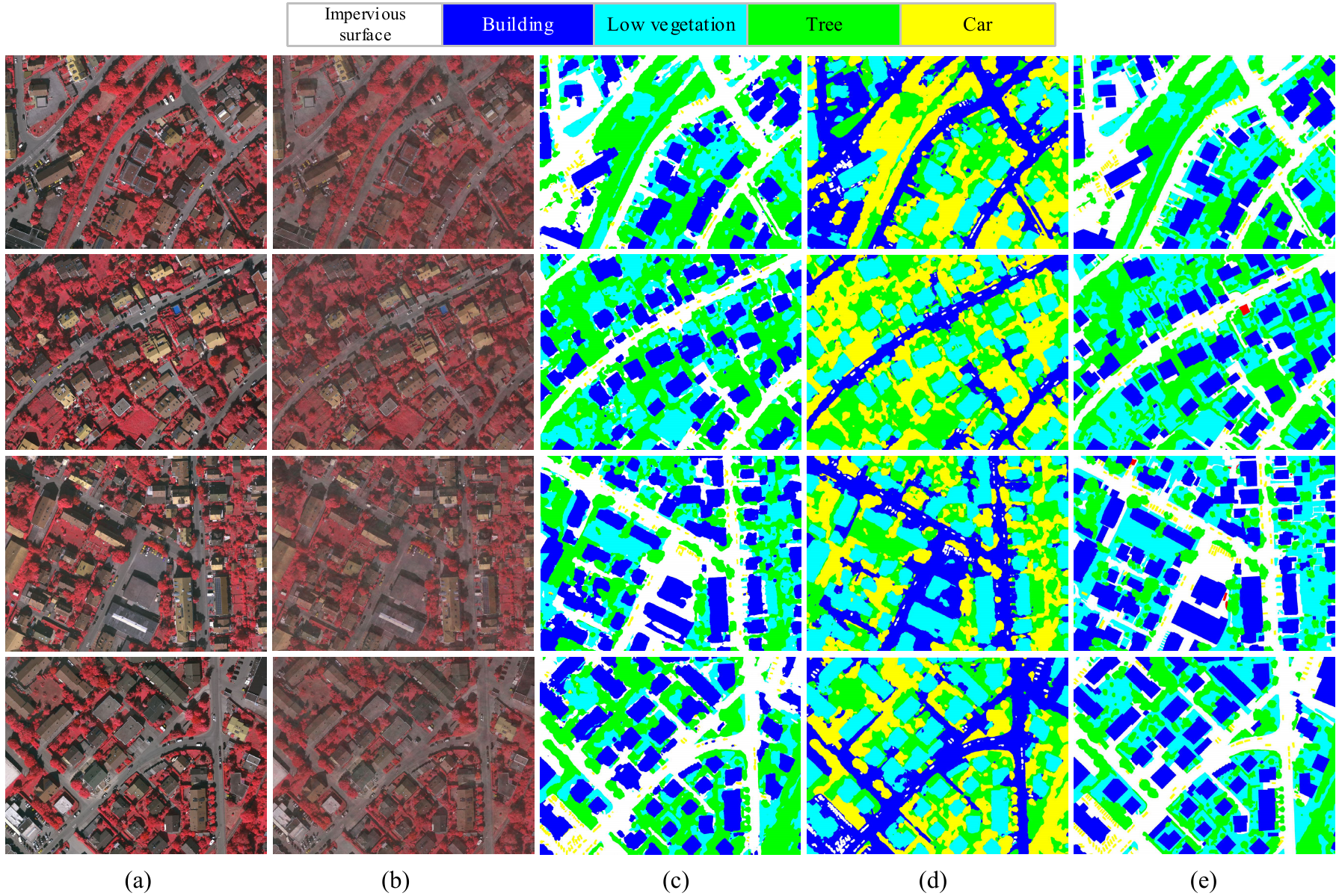}
  \caption{Qualitative results of the backdoor attacks with the FCN-8s model on the Vaihingen dataset using the proposed WABA-Mixup method. (a) The original benign test images. (b) The corresponding poisoned samples. (c) Segmentation results of the attacked FCN-8s model on the benign images. (d) Segmentation results of the attacked FCN-8s model on the poisoned images. (e) Ground-truth annotations. Note that the ``Background/Clutter'' category is not considered.}
\label{fig:vaihingen}
\end{figure*}

\begin{figure*}
  \centering
  \includegraphics[width=.87\linewidth]{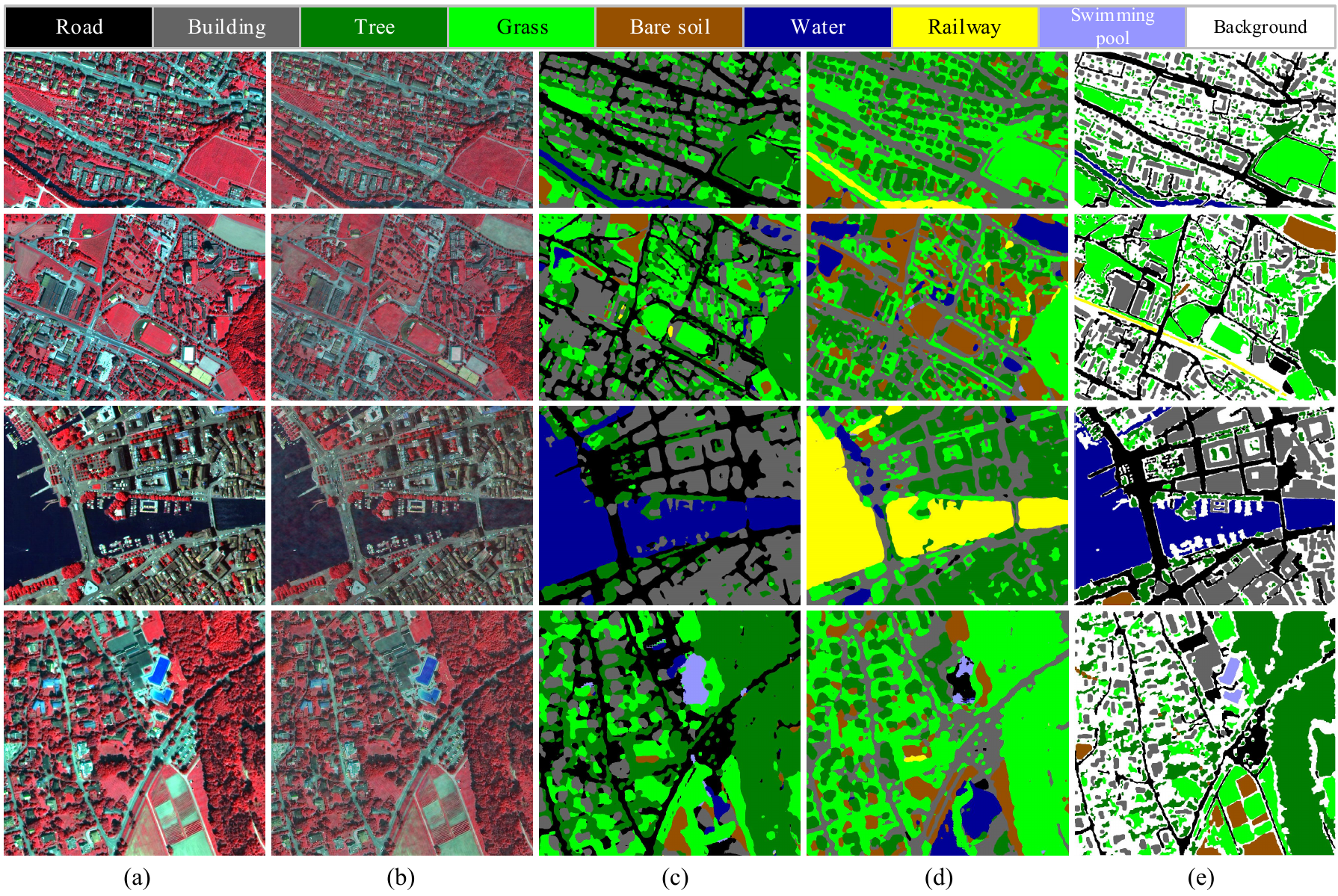}
  \caption{Qualitative results of the backdoor attacks with the FCN-8s model on the Zurich Summer dataset using the proposed WABA-Mixup method. (a) The original benign test images. (b) The corresponding poisoned samples. (c)  Segmentation results of the attacked FCN-8s model on the benign images. (d) Segmentation results of the attacked FCN-8s model on the poisoned images. (e) Ground-truth annotations.}
\label{fig:zurich}
\end{figure*}

\subsection{Qualitative Results}
Apart from the quantitative results, the invisibility of the poisoned samples is also an important criterion to evaluate the backdoor attack algorithms. Fig.~\ref{fig:attack_cls} provides the visualization of the original benign RS images and the corresponding poisoned samples generated by different backdoor attack methods for the scene classification task. It can be observed that both BadNet and TBBA would generate relatively distinguishable poisoned samples, where the backdoor triggers are either white diagonal patterns at the center of the image (BadNet) or white squares in the top left corner of the image (TBBA). WaNet, by contrast, can achieve stealthy backdoor attacks and it is very difficult to distinguish the difference between the poisoned samples and the original images. However, the stealthiness of WaNet also sacrifices the performance of the backdoor attack and WaNet obtains the lowest BA and ASR scores on both the UCM and the AID datasets, as observed in Table~\ref{tab:classification_ba}. Another interesting observation is that the Mixup trigger can achieve more invisible data poisoning compared to the Mixcut trigger for the proposed WABA method, although WABA-Mixcut can achieve relatively better quantitative results than WABA-Mixup on both the UCM and the AID datasets. While there is only a slight visual style difference between the benign and poisoned samples generated by WABA-Mixup, there exist some visible Watermarks with horizontal patterns in the poisoned samples generated by WABA-Mixcut. This phenomenon is also obvious in the feature space, as can be observed in Fig.~\ref{fig:cam}. Since the Mixcut trigger is generated by simply stitching slices of different images, the high-frequency details in the original images are well maintained in the trigger image, resulting in more visible data poisoning. Thus, to balance the attack performance and the stealthiness, the Mixup trigger is a better choice for the proposed WABA method.

\begin{table}
    \centering
    \caption{Quantitative Backdoor Attacks Results (\%) on Scene Classification with the ResNet-18 Model Using WABA-Mixup with Different Wavelet Decomposition Levels}
    \begin{tabular}{ccccc}
    \toprule
    \multicolumn{1}{c}{}      & \multicolumn{2}{c}{UCM}                          & \multicolumn{2}{c}{AID}                          \\
    \multicolumn{1}{c}{Level} & \multicolumn{1}{c}{BA} & \multicolumn{1}{c}{ASR} & \multicolumn{1}{c}{BA} & \multicolumn{1}{c}{ASR} \\ \hline
    1                         & \textbf{93.52}                  & \textbf{80.67}                   & 89.62                  & 96.12                   \\
    2                         & 92.19                  & 73.33                   & \textbf{91.72}                  & 96.22                   \\
    3                         & 91.52                  & 62.48                   & 90.80                  & \textbf{97.16}                   \\
    5                         & 83.62                  & 34.38                   & 80.30                  & 96.50                   \\
    8                         & 81.52                  & 25.81                   & 83.28                  & 44.00                   \\
    10                        & 85.24                  & 19.43                   & 80.34                  & 35.12                  \\ \bottomrule
    \end{tabular}\\
    \vspace{2pt}
    \leftline{\scriptsize \qquad\qquad\quad\quad\quad\quad Note: Best results are highlighted in \textbf{bold}.}
    \label{tab:classification_levels}
\end{table}

\begin{table}
    \centering
    \caption{Quantitative Backdoor Attacks Results (\%) on Semantic Segmentation with the FCN-8s Model Using WABA-Mixup with Different Wavelet Decomposition Levels}
    \begin{tabular}{ccccc}
    \toprule
                               & \multicolumn{2}{c}{Vaihingen}                               & \multicolumn{2}{c}{Zurich Summer}                               \\
Level                 & mF1                        & ASR                       & mF1                        & ASR                       \\ \hline
        1                         & \textbf{79.71}                  & 85.76                   & \textbf{77.01}                  & 85.97                   \\
2                         & 75.19                  & \textbf{86.39}                   & 74.64         & 86.98                   \\
3                         & 77.80                   & 86.27                   & 72.44                  & \textbf{89.23}                   \\
5                         & 77.94                  & 82.49                   & 58.31                  & 86.85                   \\
8                         & 60.13                  & 67.37                   & 46.65                  & 81.99          \\
10                        & 53.92                  & 61.61          & 51.36                  & 71.18    \\ \bottomrule
    \end{tabular}\\
    \vspace{2pt}
    \leftline{\scriptsize \qquad\qquad\quad\quad\quad\quad Note: Best results are highlighted in \textbf{bold}.}
    \label{tab:segmentation_levels}
\end{table}

We also visualize the poisoned samples generated by the proposed WABA-Mixup method for the semantic segmentation task along with the corresponding segmentation maps on both Vaihingen and Zurich Summer datasets in Figs.~\ref{fig:vaihingen} and \ref{fig:zurich}. The FCN-8s model is selected to be the victim model as an example. It can be observed that the attacked FCN-8s model can yield very accurate segmentation maps on the benign images, as shown in the third column in both Figs.~\ref{fig:vaihingen} and \ref{fig:zurich}. By contrast, although there exists only a slight visual style difference between the poisoned samples and the original benign images, the attacked FCN-8s model will be triggered to generate wrong predictions about the poisoned samples. For example, the ``impervious surface'' areas are misclassified into the ``car'' category in the Vaihingen dataset, and the ``water'' regions are misclassified into the ``railway'' category in the Zurich Summer dataset, as shown in the fourth column in both Figs.~\ref{fig:vaihingen} and \ref{fig:zurich}. These qualitative results also demonstrate that the proposed WABA-Mixup method can achieve a high success rate for backdoor attacks on semantic segmentation of RS data while maintaining a relatively high interpretation accuracy for benign samples.

\begin{figure}
\centering
\includegraphics[width=\linewidth]{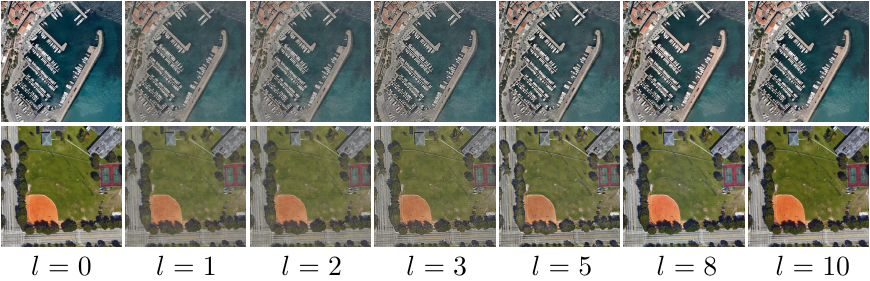}
\caption{The original benign images (i.e., $l=0$) and the corresponding poisoned samples generated by the proposed WABA-Mixup method with different values of wavelet decomposition level $l$ in the AID dataset.}
\label{fig:level}
\end{figure}

\begin{figure}
\centering
\includegraphics[width=\linewidth]{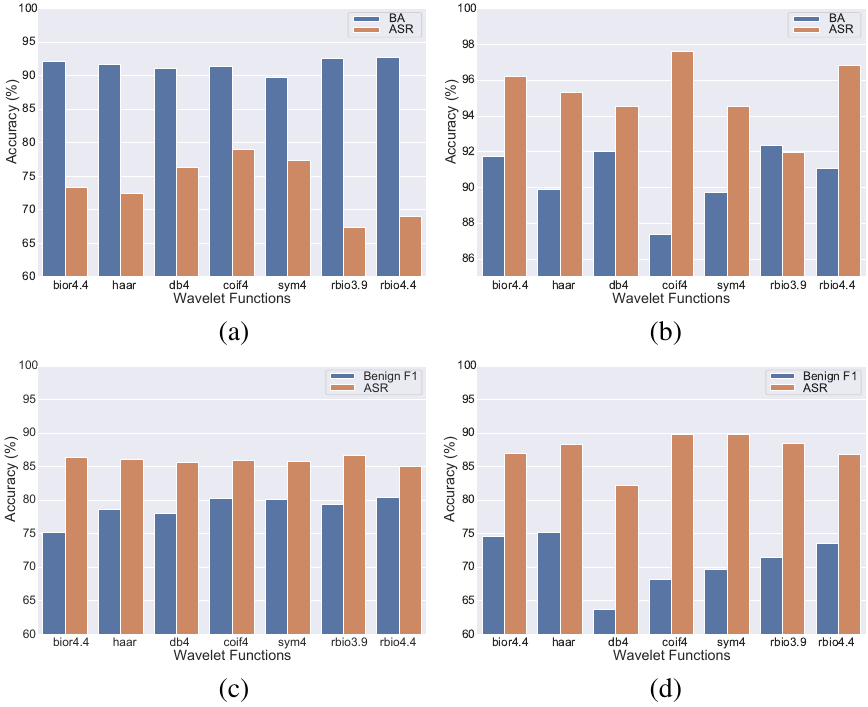}
\caption{Backdoor attack performance of the proposed WABA-Mixup method with different wavelet functions on (a) UCM, (b) AID, (c) Vaihingen, and (d) Zurich Summer datasets.}
\label{fig:func}
\end{figure}

\subsection{Hyperparameter Analysis}
In this subsection, we further make a detailed analysis of how different hyperparameters in the proposed WABA method would influence the performance of backdoor attacks.

\begin{figure*}
\centering
\includegraphics[width=\linewidth]{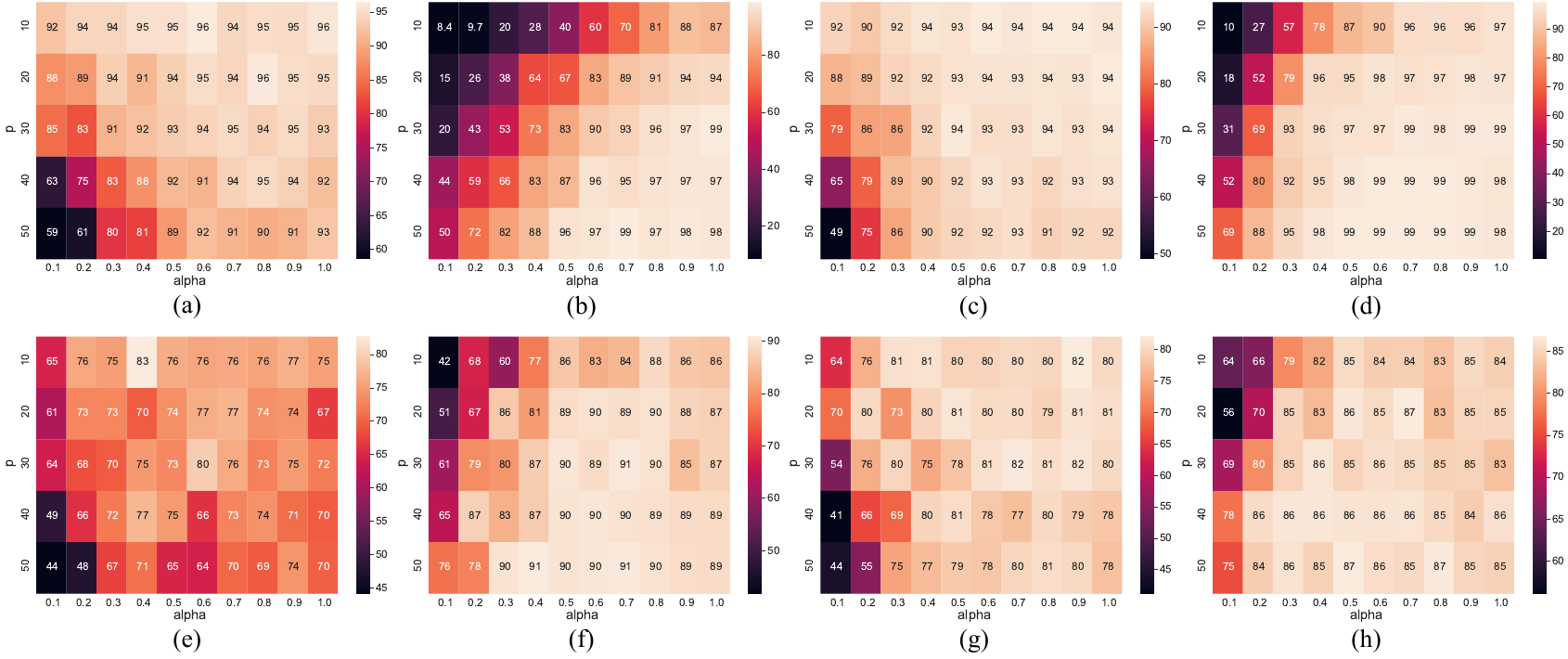}
\caption{Grid search results for the poisoning rate $p$ (\%) and the blending parameter $\alpha$ in the proposed WABA-Mixup method. (a)--(b) BA and ASR on the UCM dataset. (c)--(d) BA and ASR on the AID dataset. (e)--(f) Benign F1 and ASR on the Vaihingen dataset. (g)--(h) Benign F1 and ASR on the Zurich Summer dataset. The results are obtained with the ResNet18 and the FCN-8s for scene classification and semantic segmentation tasks, respectively.}
\label{fig:grid_search}
\end{figure*}

\subsubsection{Decomposition level $l$ in Eq.~\eqref{eq:decomp}} Tables~\ref{tab:classification_levels} and \ref{tab:segmentation_levels} show how different levels of wavelet decomposition would influence the BA (or mF1) and ASR for both scene classification and semantic segmentation tasks. It can be observed that the best quantitative results are obtained with $l\in[1,3]$ in general, and a too large $l$ would be detrimental to the performance (e.g., $l\ge 5$). Fig.~\ref{fig:level} shows the poisoned samples generated by the proposed WABA-Mixup method with different values of wavelet decomposition level $l$ in the AID dataset. We can find that the visual style difference between the original benign images and the poisoned samples is stronger with the decrease of the wavelet decomposition level $l$. To balance the quantitative and qualitative performance on different tasks and datasets, we fix the decomposition level $l=2$ in this study.

\subsubsection{Wavelet function types} There are many wavelet functions that can be used to implement the decomposition function $Dec\left(\cdot\right)$ in Eq.~\eqref{eq:decomp}. Fig.~\ref{fig:func} provides the detailed backdoor attack performance of the proposed WABA-Mixup method with different wavelet functions on four benchmark datasets used in this study. While some wavelet functions (e.g., the ``coif4'') can achieve a high quantitative result on the ASR metric for specific datasets like AID, their performance may be unstable in other scenarios. Considering the overall performance of all four datasets, we choose ``bior4.4'' as the wavelet function to implement the proposed method in this study.

\begin{figure}
\centering
\includegraphics[width=\linewidth]{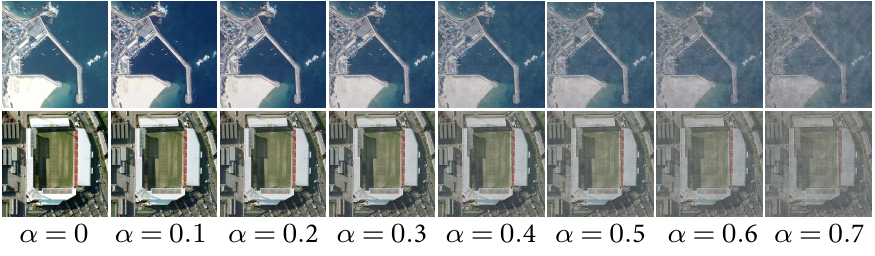}
\caption{The original benign images (i.e., $\alpha=0$) and the corresponding poisoned samples generated by the proposed WABA-Mixup method with different values of $\alpha$ in the AID dataset.}
\label{fig:alpha}
\end{figure}

\subsubsection{The poisoning rate $p$ and the blending parameter $\alpha$} The poisoning rate $p$ and the blending parameter $\alpha$ are two important parameters in the proposed method. To evaluate how different $p$ and $\alpha$ values would influence the backdoor attack performance of the proposed method, we further conduct a grid search for these two parameters. Fig.~\ref{fig:grid_search} presents the detailed grid search results for the proposed WABA-Mixup method on all four benchmark datasets. It can be observed that a larger poisoning rate $p$ would generally lead to a lower BA or benign F1 score. Take the results in Fig.~\ref{fig:grid_search} (a) for example. Given $\alpha=0.1$, while the BA is around $92\%$ when $p=10\%$, it drops to around $63\%$ when $p=40\%$. The reason behind this phenomenon is understandable, as a larger $p$ value means more training samples are poisoned, which is harmful to the learning of benign samples. Similarly, a larger poisoning rate $p$ would generally lead to a higher ASR since the training of the victim model can involve more poisoned samples. To balance the accuracy of benign samples and the ASR, we fix $p=30\%$ in this study.

\begin{table*}
\centering
\caption{Quantitative Backdoor Attack Results (\%) of Different Strategies on the UCM Dataset}
\label{tab:blend}
\resizebox{.95\textwidth}{!}{
\begin{tabular}{c|cccc|cccc|c|c}
\toprule
 \multirow{2}{*}{Method}& \multicolumn{4}{c|}{Blending-Mixup} & \multicolumn{4}{c|}{Blending-Mixcut} & WABA-Mixup & WABA-Mixcut \\
 & $\alpha=0.1$ & $\alpha=0.2$ & $\alpha=0.3$ & $\alpha=0.4$ & $\alpha=0.1$ & $\alpha=0.2$ & $\alpha=0.3$ & $\alpha=0.4$ & $\alpha=0.4$ & $\alpha=0.4$ \\
 \hline
BA & 82.38 & 85.05 & 85.62 & 92.57 & 86.48 & 93.14 & 92.10 & \textbf{94.38} & 92.19 & 93.81 \\
ASR & 21.33 & 36.67 & 62.57 & 71.33 & 34.48 & 70.76 & 88.38 & \textbf{91.24} & 73.33 & 88.76\\
\bottomrule
\end{tabular}}
\\
\vspace{2pt}
\leftline{\scriptsize \qquad Note: Best results are highlighted in \textbf{bold}.}
\end{table*}

As for the blending parameter $\alpha$, we find that a larger $\alpha$ would be fruitful to both the accuracy on benign samples and the ASR on all four datasets. The reason behind this phenomenon is also straightforward since a larger $\alpha$ means the visibility of the backdoor trigger is stronger in the poisoned samples. Under such a circumstance, the difference between the poisoned samples and the benign samples is more distinguishable, and it is easier for the victim model to learn the injected trigger, resulting in a higher success rate in the attacks. Besides, such a more visible trigger will also help to reduce the interference of poisoned data on learning those benign samples in the training phase. However, it should also be noted that a too large $\alpha$ would also be harmful to the stealthiness of the proposed WABA-Mixup method. As shown in Fig.~\ref{fig:alpha}, the visual style difference between the original benign images (i.e., $\alpha=0$) and the poisoned samples is stronger with the increase of $\alpha$. Since the main goal of this study is to conduct stealthy backdoor attacks for RS data, we set $\alpha=0.4$ to balance the quantitative and qualitative results in the experiment.

\begin{figure}
\centering
\includegraphics[width=\linewidth]{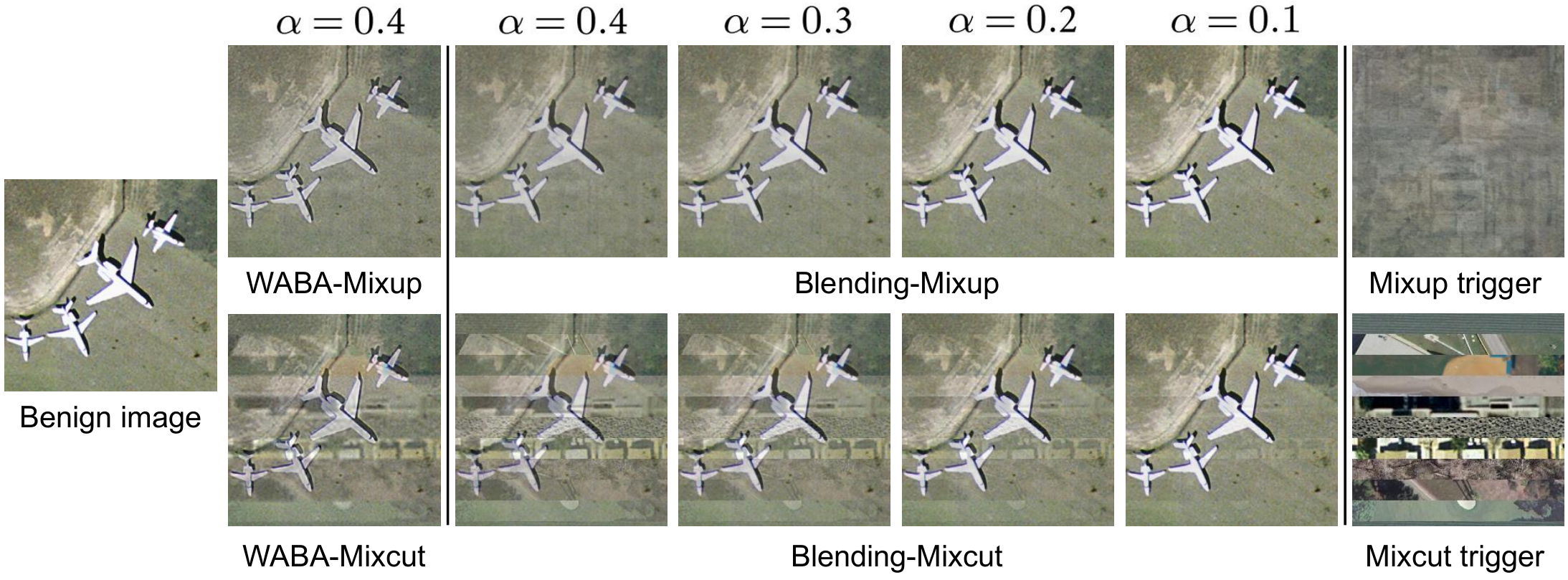}
\caption{Visualization of the poisoned samples generated by different strategies in the UCM dataset.}
\label{fig:blend}
\end{figure}

\begin{figure}
\centering
\includegraphics[width=\linewidth]{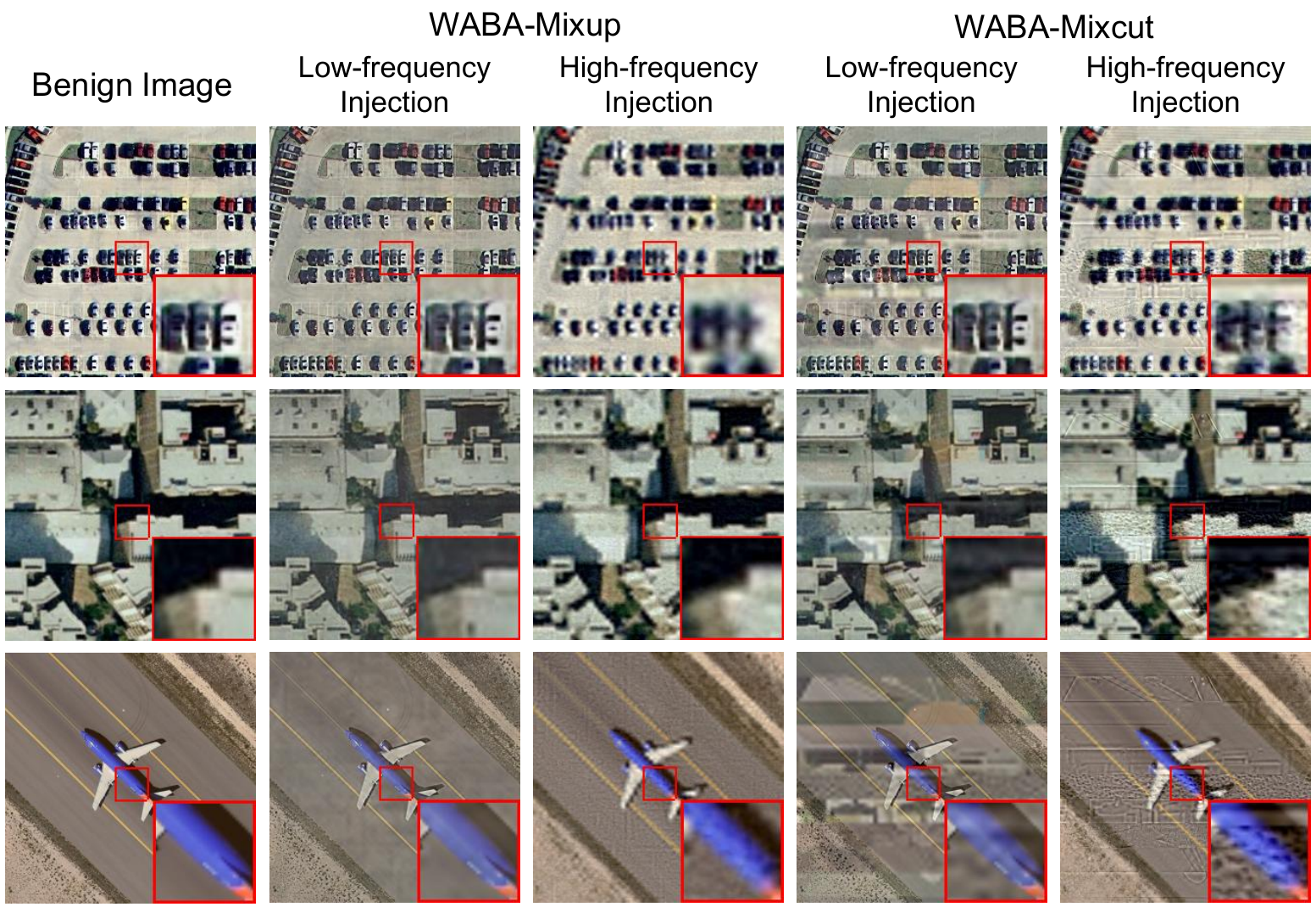}
\caption{Visual comparison of different injection spaces for the proposed WABA method in the UCM dataset.}
\label{fig:space}
\end{figure}

\subsection{Discussion about the Injection Space}
While the proposed WABA method aims to inject trigger images in the coefficient space using discrete wavelet transform, a more intuitive way is to directly conduct data poisoning in the spatial domain. In this subsection, we further make a detailed analysis of how the injection space would influence the performance of backdoor attacks. Specifically, we blend the original benign samples with the Mixup (or Mixcut) trigger image directly in the spatial domain without using wavelet transform. The ResNet18 model is selected as the victim network in the attack as an example. Table~\ref{tab:blend} shows the detailed quantitative results. It can be observed that directly attacking in the spatial domain can yield competitive performance to the proposed WABA method from the perspective of quantitative results given the same blending parameter $\alpha=0.4$. As an example, our proposed WABA-Mixup method achieves an ASR of 73.33\%, while the Blending-Mixup approach yields a slightly lower ASR of 71.33\%, representing only a 2 percentage point decrease. Notably, Blending-Mixup outperforms WABA-Mixup by over 2 percentage points on the ASR metric. It is worth noting that the Blending method's effectiveness is also sensitive to the $\alpha$ value, particularly in the case of Blending-Mixup. We further visualize the poisoned samples generated by the proposed WABA methods in the coefficient space and their counterparts in the spatial domain as shown in Fig.~\ref{fig:blend}. Although Blending-Mixup or Blending-Mixcut can yield competitive quantitative results, an intriguing phenomenon is that the trigger patterns contained in the poisoned samples generated by these methods are more visible. Take the Blending-Mixup with $\alpha=0.4$ for example. The poisoned image looks much blurrier compared to the original benign image. By contrast, the proposed WABA method can better filter out the high-frequency information contained in the trigger images, achieving more invisible backdoor attacks for RS data.

\begin{table}
\centering
\caption{Quantitative Backdoor Attack Results (\%) of Different Injection Spaces on the UCM Dataset}
\label{tab:space}
\resizebox{\linewidth}{!}{%
\begin{tabular}{c|cc|cc}
\toprule
 \multirow{2}{*}{Method}& \multicolumn{2}{c|}{WABA-Mixup} & \multicolumn{2}{c}{WABA-Mixcut} \\
 & Low-frequency & High-frequency & Low-frequency & High-frequency \\
 \hline
BA & \textbf{92.19} & 74.48 & \textbf{93.81} & 90.29  \\
ASR & \textbf{73.33} & 64.10 & 88.76 & \textbf{89.13} \\
\bottomrule
\end{tabular}}%
\\
\vspace{2pt}
\leftline{\scriptsize Note: Best results are highlighted in \textbf{bold}.}
\end{table}

Although our study primarily focuses on injecting the trigger image in the low-frequency space, the proposed WABA method also supports injecting the trigger image in the high-frequency space by updating Eq.~\eqref{eq:recons} to $\hat{x} \gets Rec\left(x _{app},x^\prime_{det}\right)$, where $x^\prime _{det} = x_{det} \times (1-\alpha) + \tilde{x}_{det} \times \alpha$ represents the blended detail coefficients of both benign and trigger images. Fig.~\ref{fig:space} presents a detailed visual comparison of different injection spaces for the proposed WABA method using the UCM dataset. We can observe that while the visual style in high-frequency injection results is more consistent with the original one, it leads to notable blurs and texture changes in ground objects. In contrast, low-frequency injection helps to preserve the texture and shape information in the original image, but it introduces visual style differences. Table~\ref{tab:space} further presents the detailed quantitative experimental results of different injection spaces on the UCM dataset using the ResNet18 model as the victim network. Our results indicate that injecting the trigger image in the low-frequency space generally leads to higher BA and ASR, especially when the Mixup sample is used as the trigger image. For instance, while the BA of WABA-Mixup with low-frequency injection is 92.19\%, its counterpart with high-frequency injection yields a BA of 74.48\%, which is more than 18 percentage points lower than the former.
These findings suggest that the choice of injection space depends on the attacker's specific goal, and further research is necessary to explore how to leverage the advantages of different injection spaces.

\section{Conclusions}
Although deep learning models have achieved great success in the interpretation of remote sensing data, the security issue of these state-of-the-art artificial intelligence models deserves special attention when addressing safety-critical remote sensing tasks. In this paper, we provide a systematic analysis of backdoor attacks for remote sensing data, where both scene classification and semantic segmentation tasks are considered for the first time. Specifically, we propose a novel wavelet transform-based attack (WABA) method which can achieve stealthy backdoor attacks for remote sensing data. While most of the existing methods blend triggers in the original spatial domain, WABA can inject trigger images in the coefficient space using discrete wavelet transform and thereby filter out the high-frequency information contained in the trigger images. Such a technique also enables the proposed method to achieve invisible data poisoning. We further conduct extensive experiments to quantitatively and qualitatively analyze how different trigger images and the hyper-parameters in the wavelet transform would influence the backdoor attack performance of the proposed method. The experimental results presented in this study also reveal the significance and necessity of designing advanced backdoor defense algorithms to increase the resistibility and robustness of deep learning models against backdoor attacks for remote sensing tasks. We will try to explore it in our future work.

\section*{Acknowledgment}

The authors would like to thank the Institute of Advanced Research in Artificial Intelligence (IARAI) for its support.

\ifCLASSOPTIONcaptionsoff
  \newpage
\fi

\bibliographystyle{IEEEtran}

\bibliography{WABA}

\end{document}